\documentclass{iopjournal}
\usepackage[numbers,square,sort&compress]{natbib}
\usepackage{xcolor}          
\usepackage{array}           
\usepackage{colortbl}        
\usepackage{lipsum}
\usepackage{amsfonts}
\usepackage{graphicx}%
\usepackage{multirow}%
\usepackage{amsmath,amssymb,amsfonts}%
\usepackage{subcaption}
\usepackage{amsthm}%
\usepackage{mathrsfs}%
\usepackage{textcomp}%
\usepackage{manyfoot}%
\usepackage{booktabs}%
\usepackage{makecell}
\usepackage{adjustbox}
\usepackage[table]{xcolor}
\usepackage{algorithm}%
\usepackage{pifont}
\usepackage{algorithmicx}%
\usepackage{algpseudocode}%
\usepackage{listings}%
\usepackage{float}
\usepackage{stfloats}
\usepackage{bm}
\usepackage{amssymb}
\usepackage{epstopdf}
\usepackage{hyperref}
\newtheorem{theorem}{Theorem}

\theoremstyle{remark}
\newtheorem{remark}{Remark}
\usepackage{ragged2e}
\justifying
\begin{document}

\articletype{Paper} 

\title{Noise-adapted Neural Operator for Robust Non-Line-of-Sight Imaging}
\author{Lianfang Wang$^1$\orcid{0009-0001-1062-2108}, Kuilin Qin$^1$\orcid{0009-0003-8205-9187}, Xueying Liu$^1$, Huibin Chang$^2$\orcid{0000-0002-8833-8940}, Yong Wang$^3$, and Yuping Duan$^{1,*}$\orcid{0000-0003-2433-3056}}

\affil{$^1$ School of Mathematical Sciences, Beijing Normal University, Beijing, 100875, People’s Republic of China}

\affil{$^2${School of Mathematical Sciences, Tianjin Normal University},
Tianjin, 300387, People’s Republic of China}

\affil{$^3$ School of Physics, Nankai University,
Tianjin, 300071, People’s Republic of China}

\affil{$^*$ Authors to whom any correspondence should be addressed.}

\email{doveduan@gmail.com}

\keywords{NLOS imaging, sparse scanning, noisy transient data, neural operator, multi-scale learning}

\begin{abstract}
\justifying
Non-line-of-sight (NLOS) reconstruction is a severely
ill-posed inverse problem, further complicated by low
photon counts, acquisition-dependent noise, and sparse
spatial sampling. Starting from Poisson--Gaussian measurement statistics, we adopt a Gaussian working model that motivates
a family of noise-dependent regularized reconstruction
maps. We propose the Noise-Adapted Neural Operator (NANO),
a noise-conditioned neural operator for learning this family of measurement-to-albedo mappings. Its numerical realization follows a lifting-update-projection architecture, combining
global--local transient lifting, complex-valued
multiscale spectral refinement, and an LCT-derived
reconstruction anchor. A pretrained noise estimator supplies a
measurement-specific descriptor that modulates
the latent updates. Measurements acquired at different spatial sampling densities are mapped to a common reference grid and reconstructed by a single trained model.
At this fixed discretization, we establish finite-depth
Lipschitz stability with respect to perturbations
in the transient input and supplied noise descriptor.
We also introduce recurrent refinement with shared
parameters and prove convergence to a unique fixed
point under a global contraction condition.
Experiments on simulated and real-world datasets
demonstrate improved reconstruction under the evaluated
noise, low-exposure, and sparse-sampling conditions
without retraining.
\end{abstract}

\section{Introduction}
\label{sec:intro}
Non-Line-of-Sight (NLOS) imaging enables the reconstruction of scenes outside the direct line of sight, with critical applications in security monitoring, autonomous driving, and medical diagnostics \cite{faccio2020non,isogawa2020optical,wu2025factorization}. In autonomous driving, NLOS facilitates the detection of occluded hazards, such as pedestrians around corners, to enhance vehicular safety \cite{scheiner2020seeing}. Similarly, in the medical domain, it allows for the non-invasive observation of deep-seated structures, including vasculature and tumors, through scattering media to improve diagnostic precision \cite{willomitzer2021fast,wan2025hdn}.

NLOS imaging detects hidden objects by capturing exceedingly faint, multi-bounce reflected light \cite{nam2021low, pei2021dynamic}. Although single-photon detectors offer extreme sensitivity, reconstructing the geometry of hidden targets remains a highly ill-posed inverse problem. The diffuse light-transport operator smooths spatial variations of the hidden scene, while reducing the number of relay-wall sampling points restricts the available spatial information and introduces poorly observed or unobservable components. When sparse scanning is further combined with short exposure times, the reduced photon counts increase the relative influence of shot noise, detector noise and ambient background. Consequently, perturbations in the measured transients may be substantially amplified during inversion, particularly under simultaneous low signal-to-noise ratios (SNRs) and incomplete spatial sampling~\cite{gupta2012reconstruction,o2018confocal}. Despite significant advances in acquisition hardware, sensor imperfections, stochastic photon counting, and sampling incompleteness continue to limit reconstruction stability. 

Existing approaches address the instability of NLOS reconstruction through different forms of regularization. Analytical methods invert approximate light-transport models using explicit spectral filtering, backprojection, or wave-propagation techniques~\cite{o2018confocal,young2020non,liu2019non}. Although computationally efficient, they commonly rely on sufficiently dense spatial sampling and may become sensitive to measurement noise and undersampling. Variational methods stabilize the inversion by incorporating handcrafted spatial priors to suppress noise and compensate for missing information~\cite{heide2014diffuse,liu2021non,yang2024efficient,ye2024plug}; however, the associated iterative optimization can be computationally demanding for high-throughput imaging. More recently, learning-based methods have replaced the regularizer, the optimization update, or the complete reconstruction operator with data-driven models~\cite{chen2019steady,xiao2024fast}. While these methods can efficiently recover complex scene structures, models trained under fixed degradation settings may generalize poorly when the noise statistics or spatial sampling patterns differ from those encountered during training. Developing a reconstruction method that remains stable across heterogeneous and non-stationary measurement conditions therefore remains an important challenge.

In this work, we propose the Noise-Adapted Neural
Operator (NANO), a noise-conditioned neural operator
for NLOS reconstruction.
Starting from Poisson--Gaussian measurement statistics,
we adopt a Gaussian working model that motivates
noise-dependent regularized reconstruction.
NANO follows a lifting--update--projection architecture,
implemented through convolutional and complex-valued
spectral processing on a fixed reference grid.
A pretrained noise-level estimation (NLE) module
provides a scalar descriptor that modulates the
latent refinement updates.
Global--local transient lifting combines information
across relay-wall locations with local temporal
features, while multiscale refinement and an
LCT-derived reconstruction anchor incorporate
learned and model-based information.
Measurements acquired at different spatial sampling
densities are mapped to the reference grid before
network evaluation, allowing a single trained model
to process these acquisition settings.

At the chosen discretization, we establish
finite-depth Lipschitz stability of the volume
reconstruction map with respect to perturbations
in the transient measurement and supplied noise
descriptor.
We also introduce recurrent reconstruction by
reusing the trained update with fixed conditioning
inputs, and prove geometric convergence to a unique
fixed point under a global contraction condition.
Experiments on simulated and real measurements
evaluate reconstruction accuracy, noise sensitivity,
and performance under reduced exposure and spatial
sampling.

To sum up, our main contributions are as follows:
\begin{itemize}
\item
We develop NANO, a noise-conditioned neural operator
that combines global--local transient lifting,
complex-valued multiscale refinement, and an
LCT-derived reconstruction anchor on a common
reference grid.

\item
We incorporate a pretrained noise estimator for
measurement-specific conditioning and establish
finite-depth Lipschitz stability with respect to
the transient input and supplied descriptor.

\item
We formulate recurrent reconstruction with shared
parameters and fixed internal depth, and establish
geometric convergence to a unique fixed point
under a global contraction condition.

\item
We evaluate NANO on simulated and real measurements,
demonstrating improved reconstruction under the
tested noise, exposure, and sampling conditions
without retraining.
\end{itemize}

\section{Preliminaries}
In this section, we begin with the mathematical formulation of the NLOS physical forward model  and then review the operator-learning framework used to motivate the reconstruction architecture.

\subsection{Forward model of NLOS}
\begin{figure}
  \centering
\includegraphics[width=0.6\linewidth]{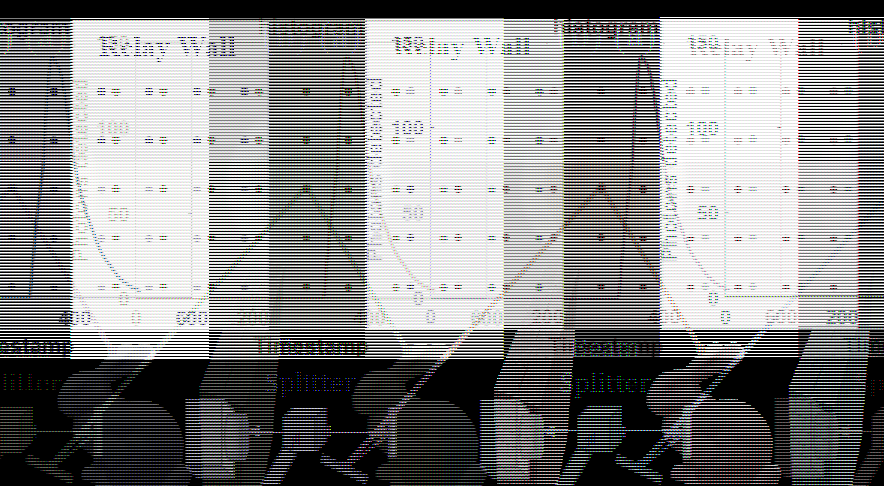}
   \vspace{-0.2cm}
   \caption{Overview of confocal imaging and measurements. {(a)} Histogram measured at the scanning points on the relay wall. {(b)} Confocal NLOS system.}
   \label{fig:Mea}
\end{figure}

Let $\Omega\subset\{\mathbf x=(x,y,z)\in\mathbb R^3:z>0\}$
denote the hidden-scene domain, with the planar relay
wall located at $z=0$.
A pulsed laser illuminates a wall point
$\mathbf x_i'=(x_i',y_i',0)$, and the returning light
is measured at $\mathbf x_d'=(x_d',y_d',0)$.
We consider the confocal configuration illustrated
in Figure~\ref{fig:Mea}, where the illumination
and detection points coincide:
$\mathbf x'=\mathbf x_i'=\mathbf x_d'$.

Under an idealized isotropic scattering model,
neglecting occlusion and additional interreflections,
the noiseless transient is represented by
\begin{equation}
\begin{aligned}
\tau(\textbf{x}^{\prime},t)=\iiint_\Omega\frac{u(\textbf{x})}{\left\|\textbf{x}^{\prime}-\textbf{x}\right\|^4}\cdot\delta\Big(2\left\|\textbf{x}^{\prime}-\textbf{x}\right\|-tc\Big)d\textbf{x},
\end{aligned}
\label{eqconnlos}
\end{equation}
where $u$ denotes the volumetric albedo,
$c$ is the speed of light, and $\delta$ is the
Dirac distribution.
Here, $t$ is the travel time after calibration
of the external propagation paths, and constant
acquisition factors are absorbed into the signal
normalization.
The delta distribution imposes the round-trip
time-of-flight constraint between the relay wall
and the hidden scene.
The factor $\|\mathbf x'-\mathbf x\|_2^{-4}$
models geometric attenuation along the outgoing
and returning paths.

NLOS reconstruction seeks to recover $u$ from
measurements of $\tau$ over the relay wall and time.
The smoothing effect of light transport and incomplete
measurement coverage make this inversion sensitive
to noise and necessitate regularization.
For the idealized confocal model, the light-cone
transform (LCT) \cite{o2018confocal} uses temporal
and depth resampling to obtain a convolutional
representation.
This structure facilitates numerical inversion
and provides model-based information for learned
reconstruction.
Departures from the assumed scattering and visibility
conditions, together with uncertainty in the instrument
response, introduce discrepancies between the model
and measured data.

\subsection{Neural operators}

Neural operators
\cite{li2020fourier,LuDeepOnet,nganyu2023deep,
azizzadenesheli2024neural,chen2024pseudo}
are parameterized models for approximating mappings
between function spaces.
In imaging inverse problems, let $\mathcal X$
denote the measurement space and $\mathcal U$
the space of unknown images or physical fields.
A learned reconstruction operator takes the form
\[
\mathcal G_\theta:\mathcal X\rightarrow\mathcal U,
\qquad
y\mapsto\mathcal G_\theta(y),
\]
where $y$ denotes the measurement and $\theta$
collects the trainable parameters. This formulation reflects the structure of imaging
inverse problems: the forward model maps an unknown
field to measurements, while reconstruction seeks
a data-to-solution map.
For an ill-posed problem, the reconstruction map
depends on the adopted prior or regularization.
Its dependence on noise statistics and acquisition
parameters can be represented by a conditional
model within a shared architecture.

A common neural-operator decomposition is
\begin{equation}
\mathcal G_\theta
=
\mathcal B_{\mathbf p}\circ\Phi\circ\mathcal B_{\mathbf l},
\label{neural-operator-structure}
\end{equation}
where $\mathcal B_{\mathbf l}$ lifts the input into
a latent representation, $\Phi$ performs learned
updates, and $\mathcal B_{\mathbf p}$ maps the
resulting representation to the output field.
The parameter dependence of these mappings is
omitted for brevity.
Conditioning variables can be incorporated into
the lifting and update mappings to represent
a parameterized family of reconstruction operators.

This formulation motivates the noise-conditioned
neural operator developed below.
Its numerical realization uses a fixed reference
grid, to which measurements from different spatial
sampling densities are mapped before reconstruction.
Generalization across these acquisition settings
is assessed experimentally and does not, by itself,
establish discretization invariance.

\section{Joint Reconstruction Model}
\label{Pre-knowledge}

In this section, we develop a noise-parameterized variational model for NLOS reconstruction. We first characterize the conditional statistics of Poisson-Gaussian measurements and discuss their transformation under resampling and Fourier representation. We then introduce an effective
Gaussian working model to obtain a tractable family of regularized reconstruction maps indexed by a scalar noise parameter.

\subsection{Mixed Noise Modeling}
\label{subsec:mixed-noise}

Under the confocal acquisition assumptions associated with \eqref{eqconnlos}, the light-cone formulation \cite{o2018confocal} uses temporal and depth resampling to obtain a convolutional representation of light transport. The discretized measurement model is
\begin{equation}
\tau=Au+\eta,
\label{forward-model}
\end{equation}
where \(u\in\mathbb R^N\) denotes the vectorized albedo volume  with \(N=N_xN_yN_z\), \(\tau\in\mathbb R^M\) is the transient measurements with \(M=TT_hT_w\), and \(\eta\in\mathbb R^M\) accounts for measurement noise. Let \(R_t\) and \(R_z\) denote the measurement-time and scene-depth resampling operators, respectively. Within the discretized convolutional approximation, the forward operator satisfies
\[
R_tA=HR_z,
\]
where \(H\) is a three-dimensional convolution operator on the resampled grid. With the boundary treatment and padding required for FFT-based convolution, the transformed model becomes
\begin{equation}
\widetilde\tau
=
\widetilde H\widetilde u+\widetilde\eta,
\label{Fouri-forward-model}
\end{equation}
where $\widetilde\tau=\mathcal F(R_t\tau)$, $\widetilde u=\mathcal F(R_z u)$, $\widetilde\eta=\mathcal F(R_t\eta)$ and \(\widetilde H\) denotes the Fourier-domain representation of \(H\), and \(\mathcal F\) is the discrete Fourier transform on the corresponding resampled grid. For simplicity, \(M\) also denotes the number of transformed measurement coefficients in the following formulation.

In photon-counting acquisition, measurement noise contains signal-dependent photon fluctuations and signal-independent readout noise. Let \(n_j\) denote the photon count in measurement bin \(j\). We adopt the conditional Poisson-Gaussian model
\begin{equation}
n_j\mid u
\sim\operatorname{Poisson}\bigl(\lambda_j(u)\bigr),
\qquad
\varepsilon_j\mid u
\sim\mathcal N(0,\sigma_g^2),
\label{eq:poisson-gaussian-components}
\end{equation}
where \(\lambda_j(u)\geq0\) is the expected photon count and \(\sigma_g^2\) is the readout-noise variance. The photon counts and readout noise are assumed to be conditionally independent within and across measurement bins. For a fixed detector gain \(\alpha>0\), assume that the nominal mean measurement satisfies \((Au)_j=\alpha\lambda_j(u)\). Then
\begin{equation}
\tau_j
=
\alpha n_j+\varepsilon_j
=
(Au)_j+\eta_j,
\label{eq:poisson-gaussian-unified}
\end{equation}
where $\eta_j:=\alpha\bigl(n_j-\lambda_j(u)\bigr)+\varepsilon_j$. 
Consequently,
\begin{equation}
\mathbb E[\eta_j\mid u]=0,
\qquad
\operatorname{Var}(\eta_j\mid u)
=
\alpha^2\lambda_j(u)+\sigma_g^2
=
\alpha(Au)_j+\sigma_g^2.
\label{eq:mixed-noise-variance}
\end{equation}
Thus, the additive representation in \eqref{forward-model} does not imply signal-independent noise. For sufficiently large expected photon counts, the Poisson component admits a Gaussian approximation, leading to
\[
\eta_j\mid u
\approx
\mathcal N\left(
0,\alpha(Au)_j+\sigma_g^2
\right).
\]
The Gaussian approximation remains signal-dependent and may be inaccurate in low-count bins. Resampling and Fourier transformation can also introduce correlations among noise coefficients. Thus, the transformed perturbations are not generally independent or identically distributed. 

To account for systematic deviations from the nominal
forward response, we introduce an effective linear
operator
\[
\widetilde H_\kappa
=
\mathcal K_\kappa(\widetilde H),
\]
where $\kappa$ collects the parameters describing
the response correction.
These parameters characterize systematic changes
in the modeled light-transport or instrument response,
whereas $\sigma$ characterizes the effective
perturbation scale.
For each fixed $\kappa$, the corrected operator
$\widetilde H_\kappa$ is linear in $\widetilde u$.
The correction map $\mathcal K_\kappa$ itself
need not be linear in $\widetilde H$.
Throughout the following variational formulation,
$\kappa$ is treated as fixed and is not jointly
estimated with $\widetilde u$ and $\sigma$.

Then the transformed measurement is represented as
\[
\widetilde\tau
=
\widetilde H_\kappa\widetilde u+e_\kappa,
\]
where $e_\kappa$ includes measurement noise and
any discrepancy not captured by the effective
operator.
Its conditional mean is small when
$\widetilde H_\kappa\widetilde u$ accurately
approximates the conditional mean measurement.
This does not, however, imply Gaussianity or
independence of the remaining perturbations. For computational tractability, we adopt the
isotropic complex Gaussian working likelihood
\begin{equation}
p(\widetilde\tau\mid\widetilde u,\sigma)
=
\frac{1}{(\pi\sigma^2)^M}
\exp\left(
-\frac{
\|\widetilde\tau-\widetilde H_\kappa\widetilde u\|_2^2
}{\sigma^2}
\right),
\label{eq:effective-gaussian-approx}
\end{equation}
where the dependence on the fixed parameter
$\kappa$ is suppressed.
The unknown scalar $\sigma>0$ denotes the effective
noise scale.
Under this working model, $\sigma^2$ is the expected
squared magnitude of each centered complex
perturbation coefficient, and its real and imaginary
components each have variance $\sigma^2/2$.

\begin{remark}
\label{prop:gaussianization}
The response correction modifies the modeled
conditional mean; it does not Gaussianize or whiten
the noise. The working likelihood additionally neglects
heteroscedasticity, spectral correlations,
residual conditional bias, and the conjugate
symmetry inherited from real-valued measurements.
It therefore provides a tractable surrogate
rather than an exact reformulation of the
Poisson--Gaussian model.
\end{remark}

\subsection{Noise-Parameterized NLOS Formulation}
\label{subsec:noise-parameterized}

Using the Gaussian working likelihood in \eqref{eq:effective-gaussian-approx}, we adopt the isotropic Gaussian prior
\[
\widetilde u\sim\mathcal{CN}(0,\gamma^2 I_N),
\]
where \(\gamma>0\) is fixed and \(\gamma^2\) denotes the prior variance per complex coefficient. This prior induces a quadratic penalty on the
transformed reconstruction. Combining the negative log-likelihood with this penalty and omitting terms independent of \((\widetilde u,\sigma)\) gives
\begin{equation}
\min_{\widetilde u\in\mathbb C^N,\;\sigma\geq\sigma_{\min}} 
\Big\{
\frac{
\|\widetilde H_\kappa\widetilde u-\widetilde\tau\|_2^2
}{\sigma^2}
+ M\log\sigma^2 + \frac{1}{\gamma^2}\|\widetilde u\|_2^2\Big\},
\label{LSE-setting}
\end{equation}
where \(\sigma>0\) is the effective noise scale and \(\sigma_{\min}>0\) is a prescribed lower bound. The constraint prevents variance collapse: if an exact fit is attainable, the unconstrained objective tends to \(-\infty\) as \(\sigma\to0\). For fixed \(\widetilde H_\kappa\) and \(\gamma>0\),
the constrained finite-dimensional objective is continuous and has compact sublevel sets, and hence admits a minimizer.

For fixed \(\widetilde u\), the unique minimizer with respect to the noise scale is
\begin{equation}
\sigma^\ast(\widetilde u)
= \max\Big\{ \sigma_{\min},
\frac{\|\widetilde H_\kappa\widetilde u-\widetilde\tau\|_2}{\sqrt M}
\Big\}.
\label{eq:noise_variance_update}
\end{equation}
Thus, the conditional estimate is the root mean squared data discrepancy, bounded below by \(\sigma_{\min}\).

Conversely, for fixed \(\sigma\geq\sigma_{\min}\), the logarithmic term is independent of \(\widetilde u\). Multiplying the remaining objective by \(\sigma^2\) yields the equivalent Tikhonov problem
\begin{equation}
\min_{\widetilde u\in\mathbb C^N}
\left\{
\|\widetilde H_\kappa\widetilde u-\widetilde\tau\|_2^2
+
s_\sigma\|\widetilde u\|_2^2
\right\},
\qquad
s_\sigma:=\frac{\sigma^2}{\gamma^2}.
\label{eq:fixed-noise-inverse}
\end{equation}
Its first-order optimality condition is
\begin{equation}
\left(
\widetilde H_\kappa^\ast\widetilde H_\kappa
+
s_\sigma I_N
\right)\widetilde u
=
\widetilde H_\kappa^\ast\widetilde\tau,
\label{normal-equation}
\end{equation}
where \(\widetilde H_\kappa^\ast\) denotes the adjoint of \(\widetilde H_\kappa\). Since \(s_\sigma\geq\sigma_{\min}^2/\gamma^2>0\), the normal-equation matrix is Hermitian positive definite. The conditional reconstruction therefore has the unique solution
\[
\widetilde u^\ast
=
\left(
\widetilde H_\kappa^\ast\widetilde H_\kappa
+
s_\sigma I_N
\right)^{-1}
\widetilde H_\kappa^\ast\widetilde\tau.
\]

For a fixed effective forward operator,
\eqref{eq:fixed-noise-inverse} defines a family of
regularized reconstruction maps parameterized by
$s_\sigma=\sigma^2/\gamma^2$.
This parameter controls the balance between data
fidelity and prior regularization.
To avoid coupling noise estimation to the evolving
reconstruction residual, we adopt a plug-in strategy:
the noise scale is estimated once from the measurement
and used to construct a scalar conditioning descriptor,
which remains fixed throughout reconstruction.
This motivates the noise-conditioned reconstruction
map developed in the next section, where the
descriptor's relation to the theoretical regularization
parameter is specified.



\section{Neural-operator-based Reconstruction}\label{network-architecture}
In this section, we realize the plug-in formulation using a noise-level estimator followed by a noise-adapted neural operator. Given a measured transient \(\tau\), the estimator produces \({\sigma}\), from which the scalar conditioning descriptor \(s_{{\sigma}}\) is constructed. The neural operator then maps \((\tau,s_{{\sigma}})\) to an estimate of \({u}\). 

\begin{figure}
  \centering
   \includegraphics[width=0.9\linewidth]{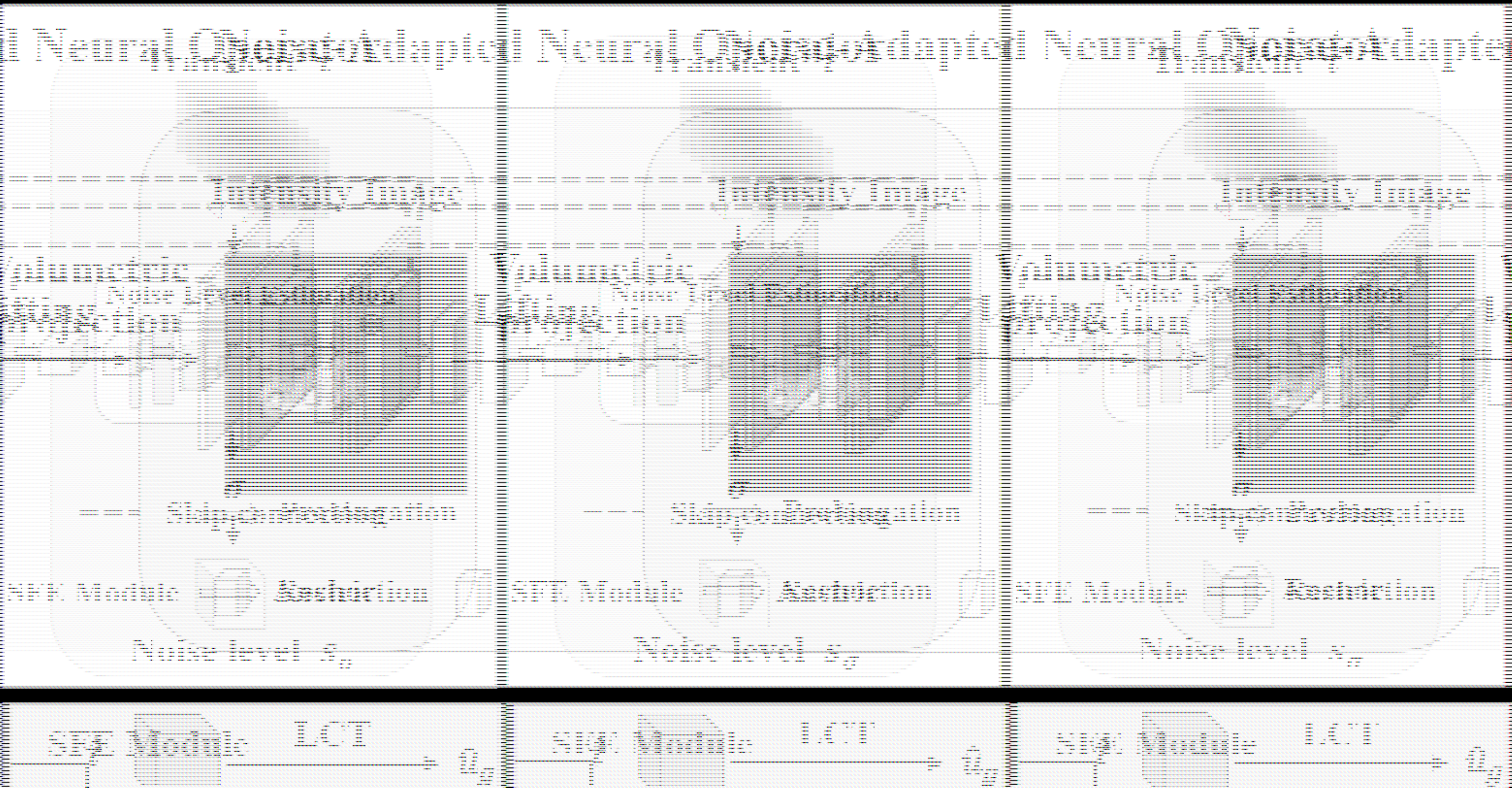}
   \vspace{-0.1cm}
   \caption{Flowchart of the proposed Noise-Adapted Neural Operator.}
   \label{fig:model}
\end{figure}

\subsection{Plug-in Estimation for the \(\sigma\)-Subproblem}
The conditional noise-scale update in \eqref{eq:noise_variance_update} depends on the unknown reconstruction. To avoid alternating between reconstruction and noise estimation, we use a data-driven plug-in
estimator. The noise-level estimation (NLE) module learns to predict a noise scale from the observed transient, using supervised calibration data that include variability in both the clean scene and the noise magnitude.

We calibrate the supervised estimator using measurements with known noise scales. For a prescribed noise standard deviation \(\sigma_{\mathrm{gt}}\), we generate
\[
\tau=\tau_0+n,
\qquad
n\sim\mathcal N(0,\sigma_{\mathrm{gt}}^2I_M),
\]
where \(\tau_0\) denotes the corresponding clean transient and \(n\) is independent of \(\tau_0\). Consequently,
\[
\sigma_{\mathrm{gt}}^2
=
\frac{1}{M}\mathbb E[\|n\|_2^2]
=
\frac{1}{M}
\mathbb E\left[
\|\tau-\tau_0\|_2^2
\,\middle|\,\tau_0
\right].
\]
Thus, \(\sigma_{\mathrm{gt}}\)  is the population noise scale associated with the ideal residual-based minimizer \eqref{eq:noise_variance_update}. The NLE module learns to infer this reference scale directly from the measurement, thereby providing a plug-in approximation to the $\sigma$-subproblem without requiring the unknown reconstruction or
per-measurement optimization. Because spatial transformations and resampling operators generally alter noise energy, we employ the predicted measurement-domain scale as a calibrated surrogate for the transformed-domain update in \eqref{eq:noise_variance_update}.

We implement the NLE module using a 3D U-Net encoder-decoder architecture
\cite{dong2019deep,li2025curvpnp}. The encoder extracts transient features through successive convolutions and strided downsampling, while the output
branches predict a scalar noise scale and an auxiliary spatial noise map:
\[
(\sigma,D_n)
=
\operatorname{U\text{-}Net}_{\mathrm{3D}}
(\tau;\theta_{\mathrm{NLE}}),
\]
where \(\sigma\) denotes the predicted noise standard deviation in this subsection, and \(D_n\) is supervised by the corresponding spatial target \(D_{\mathrm{gt}}\). To establish a physically interpretable reference scale
for \(\sigma\), the NLE module is pretrained on the synthetic dataset from \cite{chen2020learned} using zero-mean Gaussian perturbations with varying standard deviations. The NLE module is optimized by
\[
\mathcal L_{\mathrm{NLE}}
=
|\sigma-\sigma_{\mathrm{gt}}|
+
\beta\|D_n-D_{\mathrm{gt}}\|_1,
\]
where \(\beta>0\) is a balancing hyperparameter, and \(\sigma_{\mathrm{gt}}\) and \(D_{\mathrm{gt}}\) denote the ground-truth noise standard deviation and auxiliary noise map, respectively. After pretraining, the module's weights are frozen to provide consistent noise descriptors for the subsequent reconstruction network.

In the numerical implementation, the prior variance
$\gamma^2$ is not estimated explicitly.
Instead, we construct a dimensionless conditioning
descriptor from the estimated noise power and the
observed transient power:
\begin{equation}
s_\sigma
=
\frac{P_\sigma}{P_\tau} \quad \mbox{with}~~
P_\tau=\frac{1}{M}\|\tau\|_2^2>0~ \mbox{and}~
~
P_\sigma=\sigma^2,
\label{eq:noiselevel}
\end{equation}
where $\sigma$ is obtained by the NLE module. For notational simplicity, $s_\sigma$ denotes this empirical descriptor in the subsequent network description and serves as a conditioning surrogate for the theoretical regularization parameter $\sigma^2/\gamma^2$.  It is supplied to $\mathcal G_\sigma$ to adapt the reconstruction to the estimated noise condition.

\subsection{Noise-adapted neural operator for the $\tilde u$-Subproblem}\label{network}
In this subsection, we describe NANO reconstruction
architecture.
Throughout the numerical formulation,
\(\mathcal X\) and \(\mathcal U\) denote the
finite-dimensional measurement and reconstruction
spaces associated with the reference discretization.
The discretization index is omitted for brevity.

Given the noise descriptor \(s_\sigma\),  NANO parameterizes a
noise-indexed family of reconstruction maps
\begin{equation}
u=\mathcal G_\sigma(\tau,s_\sigma;\theta),
\qquad
\mathcal G_\sigma:
\mathcal X\times[0,\infty)\rightarrow\mathcal U,
\label{eq:iter_form}
\end{equation}
where \(\theta\) collects the trainable reconstruction
parameters.
As illustrated in Figure~\ref{fig:model}, the architecture
comprises a lifting map \(\mathcal B_{\mathbf l}\),
a latent refinement module \(\Phi\), and a volume
projection map \(\mathcal B_{\mathbf p}\).

\begin{figure}
  \centering
   \includegraphics[width=0.65\linewidth]{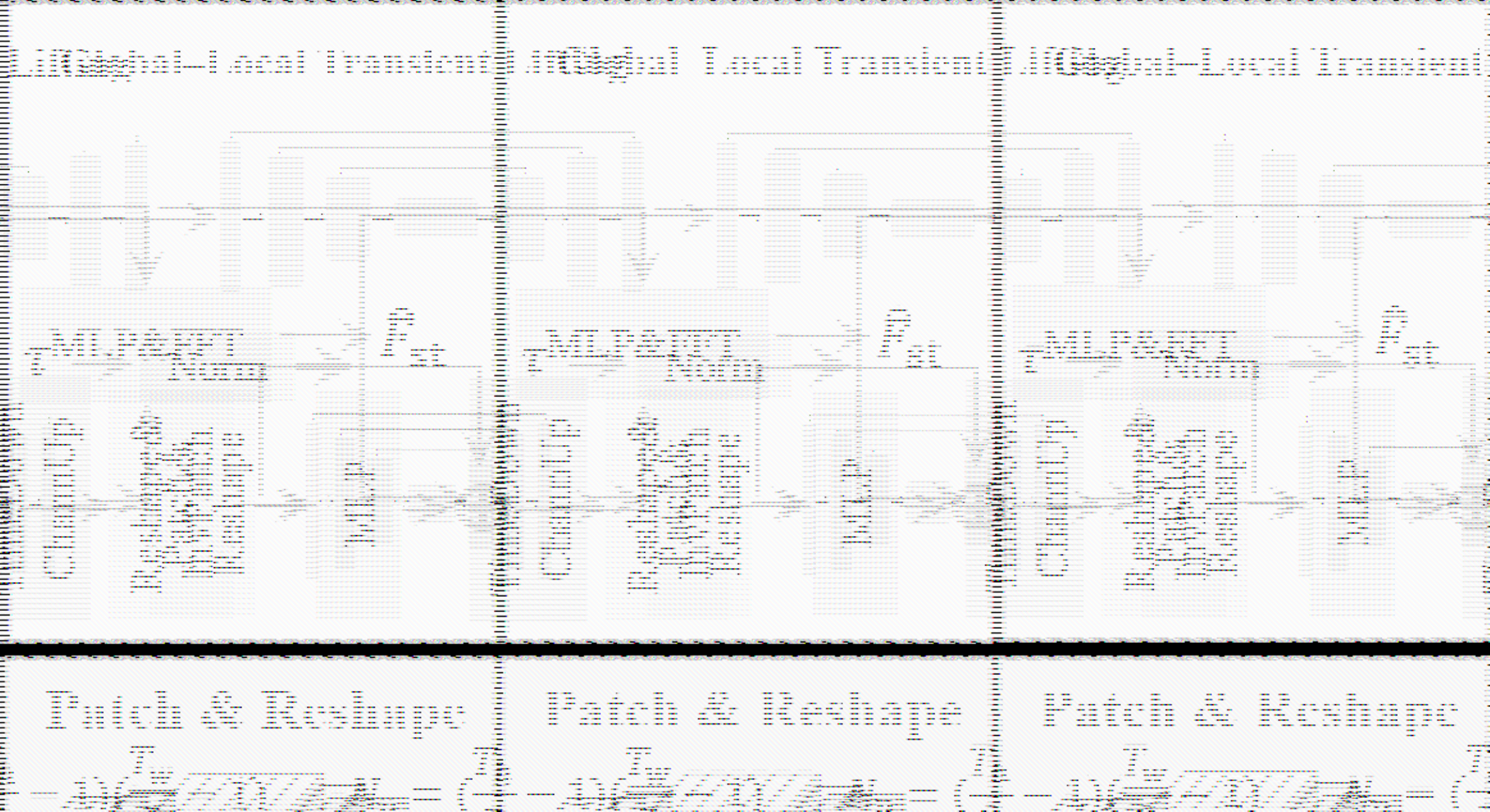}
   \vspace{-0.1cm}
   \caption{Illustration of the proposed Global–Local Transient Lifting module.}
   \label{fig:GLTL}
\end{figure}

\subsubsection{Noise-level and transient lifting}
The lifting operator \(\mathcal B_{\mathbf l}\) maps the transient measurement and the noise descriptor into a joint latent space:
\[
(\widehat\tau,\widehat s_\sigma)
=
\mathcal B_{\mathbf l}(\tau,s_\sigma)
\in\mathcal H,
\qquad
\mathcal H=\mathcal H_\tau\oplus\mathcal H_s,
\]
where \(\mathcal H_\tau\) and \(\mathcal H_s\) denote the latent measurement and noise-conditioning spaces, respectively. The space \(\mathcal H\) is equipped with the canonical Hilbert product norm.

\textbf{Noise-level lifting.}
We first broadcast the scalar descriptor $s_\sigma$ over the spatial and temporal dimensions of the transient measurement, then embed it into the latent conditioning space using a multilayer perceptron
\[
\widehat s_\sigma
=
\operatorname{MLP}(s_\sigma\mathbf{1};\theta)
\in\mathcal H_s.
\]
Through learned affine transformations and nonlinear activations, the MLP produces a tensor embedding that serves as conditioning information for the
subsequent latent refinement steps.

\textbf{Transient lifting.}
Transient measurements encode the ToF signatures of propagating light pulses across relay-wall locations. We introduce the Global-Local Transient Lifting (GL-TL) module, illustrated in Figure~\ref{fig:GLTL},
to extract complementary global and local features directly from these measurements. The input tensor is $\tau\in\mathbb R^{B\times C\times T\times T_h\times T_w}$, where \(B\), \(C\), and \(T\) denote the batch size, number of channels, and number of temporal bins, and \(T_h,T_w\) denote the spatial sampling dimensions of the relay wall.

In \textbf{Global Transient Lifting}, a 3D U-Net aggregates information across spatial locations and temporal bins. Because measurements from different detection points provide complementary views of the hidden scene,
this branch extracts features that capture their joint spatiotemporal structure:
\[
F_g = \operatorname{U\text{-}Net}_{\mathrm{3D}}(\tau;\theta)
\in \mathbb R^{B\times4C\times T\times\frac{T_h}{2}\times\frac{T_w}{2}}.
\]

In \textbf{Local Transient Lifting}, we process neighboring transient signals to capture variations in arrival times and amplitudes associated with different propagation paths. Residual convolutions first downsample the input and expand its channel dimension, yielding \(\tau_\downarrow\).
For each selected spatial center \((i,j)\), a patch-extraction operator collects a \(3\times3\) neighborhood at each temporal index:
\[
\bigl[\mathcal P(\tau_\downarrow)\bigr]_
{b,c,t,(i,j),k,l}
=
(\tau_\downarrow)_{b,c,t,i+k,j+l},
\qquad
k,l\in\{-1,0,1\}.
\]
The extraction stride and boundary treatment determine the set of patch centers. A reshaping operator \(\mathcal R\) combines the neighborhood and channel dimensions while preserving the temporal axis. A 1D U-Net then processes the temporal signals within each patch:
\[
F_p=\operatorname{U\text{-}Net}_{\mathrm{1D}}
\bigl(\mathcal R(\mathcal P(\tau_\downarrow));\theta\bigr).
\]
The patch features are remapped to the output grid of the global branch and processed by a residual MLP to obtain $F_l\in
\mathbb R^{
B\times4C\times T\times\frac{T_h}{2}\times\frac{T_w}{2}
}$. 

Finally, the global and local features are concatenated along the channel dimension and fused through a convolutional layer:
\[
F_{\mathrm{st}}
=
\operatorname{Conv}\bigl((F_l,F_g);\theta\bigr)
\in
\mathbb R^{
B\times4C\times T\times\frac{T_h}{2}\times\frac{T_w}{2}
},
\]
where \((F_l,F_g)\) denotes channel concatenation. A Fourier transform along the temporal and spatial axes converts the fused features into the latent measurement representation:
\[
\widehat\tau
:=
\operatorname{FFT}(F_{\mathrm{st}}).
\]
The resulting \(\widehat\tau\) is supplied to the subsequent complex neural operator.

\subsubsection{Complex Neural Operator} 
The latent evolution operator $\Phi$ performs $K$
learned refinement steps in the complex domain,
inspired by gradient descent for the regularized
objective in \eqref{eq:fixed-noise-inverse}.
These steps are distributed across the scales.
Within each scale, the measurement-derived conditioning
feature and noise embedding remain fixed while the
latent state is updated.
Suppressing the scale index, the refinement is written as
\begin{equation}
\widehat u^{k+1}
=
\mathcal A_{\sigma,k}
(\widehat u^k;\widehat \tau,\widehat s_\sigma),
\label{phi_composition}
\end{equation}
starting from the initial latent state at that scale.
Each refinement operator takes the form
\begin{equation}
\mathcal A_{\sigma,k}
(w;\widehat \tau,\widehat s_\sigma)
:=
w+\xi_k(\theta)
\operatorname{SConv}_{\mathbb C,k}
\bigl(\widetilde H_\kappa^*\widehat\tau+
\mathcal S_{\sigma,k}(w,\widehat s_\sigma;\theta);
\theta
\bigr),
\label{fixed-point}
\end{equation}
where the coefficient $\xi_k(\theta)\in(0,1)$ is a learnable
relaxation parameter and $\mathcal S_{\sigma,k}$
provides noise-dependent state modulation.
The spectrally normalized complex-valued block
$\operatorname{SConv}_{\mathbb C,k}$ produces a
correction with the same dimensions as $w$.
The parameter collection $\theta$ includes the learned
weights, biases, and relaxation coefficients;
its occurrence in $\mathcal A_{\sigma,k}$ is
suppressed for brevity.

To process complex feature maps, let \(\mathcal W=\mathcal M+i\mathcal N\) denote a complex kernel and \(h=x+iy\) a complex state. The linear part of a complex convolution is
\[
\operatorname{Conv}_{\mathbb C}(h)
=
(\mathcal M*x-\mathcal N*y)
+i(\mathcal M*y+\mathcal N*x),
\]
where \(*\) denotes convolution over the feature grid.  We use the component-wise activation
\[
\varphi_{\mathbb C}(h)
=
\varphi(\operatorname{Re}h)
+i\varphi(\operatorname{Im}h).
\]
If \(\varphi\) is \(L_\varphi\)-Lipschitz, then \(\varphi_{\mathbb C}\) is also \(L_\varphi\)-Lipschitz under the real-imaginary product norm. These layerwise properties support the perturbation bounds developed in Section~\ref{neural-operator-iteration}.

\textbf{Noise-level-induced operator.} Within the iterative framework \eqref{fixed-point}, the operator \(\mathcal S_{\sigma,k}\) modulates
the latent state using coefficients generated from the noise embedding \(\widehat s_\sigma\):
\[
\mathcal S_{\sigma,k}
(\widehat u^k,\widehat s_\sigma;\theta)
=
\operatorname{Conv}_{\mathbb C}
\Bigl(
\varphi_{\mathbb C}
\bigl(
\operatorname{Conv}_{\mathbb C}
(\widehat s_\sigma;\theta)
\bigr);
\theta
\Bigr)
\odot\widehat u^k,
\]
where \(\odot\) denotes elementwise multiplication with the modulation coefficients broadcast over the appropriate feature dimensions. The two complex convolutional layers have distinct,
step-specific parameters included in \(\theta\). Together with the intervening activation, they map the noise embedding to coefficients that scale and phase-modulate the complex latent features. Combined with the measurement-derived feature \(\widehat u_g\) in \eqref{fixed-point}, this mechanism conditions each refinement update
on both the measurement and the estimated noise level.

{\bf{Multi-scale hierarchical framework.}}
The latent operator \(\Phi\) uses a multiscale architecture to combine coarse-scale structural information with finer-scale features. The latent measurement \(\widehat\tau\) and noise embedding \(\widehat s_\sigma\) provide conditioning information at the corresponding scales.

The refinement is distributed across \(J\) scales. At scale \(j\in\{1,\ldots,J\}\), the model performs \(n_j\) updates, with \(\sum_{j=1}^J n_j=K\). For \(k=0,\ldots,n_j-1\), each update combines
noise-dependent state modulation with a learned spectral correction:
\begin{equation}
\widehat u^{j,k+1}
=
\widehat u^{j,k}
+
\xi^{j,k}(\theta)\,
\operatorname{SConv}_{\mathbb C}
\Bigl(
\widehat u_g^j
+
\mathcal S_{\sigma,k}^j
(\widehat u^{j,k},\widehat s_\sigma^j;\theta);
\theta
\Bigr).
\label{single}
\end{equation}
where $\widehat u_g^j$ denotes the LCT-based approximation of $\widetilde H_\kappa^*\widehat\tau^j$ at scale \(j\) and $\widehat s_\sigma^j$ is the noise embedding represented in the corresponding feature dimensions.
At the finest scale, a fixed LCT-based transformation
of the learned transient representation, followed by
channel matching, produces $\widehat u_g^1$ and
initializes $\widehat u^{1,0}=\widehat u_g^1$.
Coarser-scale conditioning features are obtained
by restriction and remain fixed during refinement
at each scale.
The operator $\mathcal S_{\sigma,k}^j$ follows
the modulation form defined above.
The spectral block learns the update direction,
while the learnable coefficient
$\xi^{j,k}(\theta)\in(0,1)$ scales its magnitude.

Complex convolutions with stride \(2\) transfer the state and measurement-derived features to coarser scales:
\[
\left(
\widehat u^{j+1,0},
\widehat\tau^{j+1},
\widehat u_g^{j+1}
\right)
=
\mathcal R_{\mathbb C}^{j+1}
\left(
\widehat u^{j,n_j},
\widehat\tau^j,
\widehat u_g^j;\theta
\right),
\qquad j=1,\ldots,J-1,
\]
where \(\widehat\tau^j\) denotes the latent measurement feature at scale \(j\) with \(\widehat\tau^1=\widehat\tau\). The refinement in \eqref{single} is then applied at the next scale. Finally, the multiscale outputs are aggregated through prolongation and skip connections to produce \(\widehat u^K\), which denotes the output of the complete latent hierarchy.

\textbf{Stochastic frequency encoding (SFE).}
At the coarsest scale, we combine spectral components
using Gaussian random weights and process the resulting
representation through a complex MLP:
\begin{equation}
v=
\operatorname{MLP}_{\mathbb C}
\Big(
\sum_{l=1}^{N_B}
M_l\,\widehat u_l^{J,n_J};\theta
\Big),
\label{cco}
\end{equation}
where \(\widehat u_l^{J,n_J}\) denotes the feature associated with the \(l\)-th frequency component, \(M_l\) is its random linear weighting map and \(N_B\) is the number of components included in the aggregation.  The complex MLP transforms the aggregated representation into the feature dimensions required by the subsequent multiscale aggregation, where \(v\) replaces the coarsest-scale feature.

This encoding combines information from different frequency components through random weighting and learned nonlinear processing. Its contribution to reconstruction is evaluated through the ablation experiments. The stability and fixed-point analyses assume a fixed realization of the random weights.

{\bf{Physics-anchored reconstruction.}}
To preserve structural details from the model-based estimate, the learned spectral features are fused with the LCT reconstruction anchor using a learnable weight:
\begin{equation}
\widetilde u^K
=
\mu(\theta)\widetilde u_g
+
\bigl(1-\mu(\theta)\bigr)
\operatorname{Conv}_{\mathbb C}(\widehat u^K;\theta),
\label{eq:physics_anchored_reconstruction}
\end{equation}
where $\widehat{u}_g$ denotes the LCT-based reconstruction of $\widehat\tau$, matched to the grid and channel dimensions of the modulation output. The coefficient \(\mu(\theta)\in(0,1)\) determines the relative contribution of the two estimates. It is learned during training and held fixed during inference.

\begin{remark}
For fixed inputs and trained parameters, the physics-anchored output is the unique minimizer of
\[
\widetilde u^K
=
\operatorname*{arg\,min}_{\widetilde u}
\Big\{
\left\|
\widetilde u
-
\operatorname{Conv}_{\mathbb C}(\widehat u^K;\theta)
\right\|_2^2
+
\frac{\mu(\theta)}{1-\mu(\theta)}
\|\widetilde u-\widetilde u_g\|_2^2
\Big\}.
\]
Thus, the fusion admits a quadratic regularization interpretation in the spectral reconstruction space, with \(\mu(\theta)/(1-\mu(\theta))\) controlling the strength of attraction toward the LCT anchor. 
\end{remark}

\subsubsection{Volumetric albedo projection}
The reconstructed volume is obtained by applying the inverse Fourier transform and inverse depth resampling $u=R_z^{-1}\operatorname{Re}\big(\mathcal F^{-1}(\widetilde u^K)\big)$, where \(\mathcal F^{-1}\) denotes the inverse FFT, $\operatorname{Re}(\cdot)$ denotes the element-wise real-part
projection and \(R_z^{-1}\) denotes the mapping back to the original depth grid. Following the use of image and depth supervision in \cite{chen2019steady}, we train the reconstruction using projected albedo and depth maps.
For a real-valued reconstructed volume \(u\), the initial albedo and depth maps are
\[
\overline I(x,y)
=
\max_j u(x,y,z_j),
\qquad
\overline D(x,y)
=
z_{j^\ast(x,y)},
\qquad
j^\ast(x,y)\in
\operatorname*{arg\,max}_j u(x,y,z_j).
\]
A fixed tie-breaking rule is used when multiple depth bins attain the maximum. The two maps are then refined by separate 2D networks comprising residual convolutional blocks:
\[
I=\operatorname{Conv}(\overline I;\theta_I),
\qquad
D=\operatorname{Conv}(\overline D;\theta_D),
\]
where \(\theta_I\) and \(\theta_D\) denote the parameters of the albedo and depth refinement networks, respectively.

\subsection{Loss function}
The reconstruction network is trained using albedo and depth supervision together with total variation (TV) regularization of the volume:
\[
\mathcal L
=
\frac{1}{HW}\|I-I_{\mathrm{gt}}\|_2^2
+
\frac{\lambda_{\mathrm d}}{HW}
\|D-D_{\mathrm{gt}}\|_1
+
\lambda_{\mathrm{tv}}\mathcal L_{\mathrm{TV}}(u),
\]
where \(H,W\) are the output image dimensions, \(I_{\mathrm{gt}}\) and \(D_{\mathrm{gt}}\) are the reference albedo and depth maps, and
\(\lambda_{\mathrm d},\lambda_{\mathrm{tv}}\geq0\) balance the loss terms. The first two terms measure the mean squared albedo error and mean absolute depth error, respectively. The TV term penalizes spatial variation in the reconstructed volume to discourage spurious oscillations.

\section{Stability and Recurrent Reconstruction}
\label{neural-operator-iteration}

We establish the stability of the finite-depth reconstruction
map and give a sufficient condition for convergence of its
recurrent extension.

\subsection{Stability Analysis}

We consider the volume reconstruction map \(\mathcal G_\sigma(\tau,s_\sigma;\theta)\) at a fixed discretization, with trained parameters and SFE weights held fixed. The measurement \(\tau\) and noise descriptor \(s_\sigma\) are treated as independent input arguments. The analysis excludes the noise-estimation module and hard argmax depth extraction.

Complex feature spaces are equipped with the real-imaginary Euclidean product norm. All affine and fixed linear mappings are bounded, and the activation functions are Lipschitz continuous. We assume that initialization, conditioning, and real-valued output extraction are Lipschitz on bounded sets and preserve boundedness. Elementwise modulation is Lipschitz on bounded feature sets, so these properties extend through the finite
computation graph. Spectral normalization provides additional control of the convolutional layer norms \cite{miyato2018spectral}. We write \(\|\cdot\|\) for feature norms and their induced operator norms, and \(|\cdot|\) for scalar absolute values. Let \(\mathcal D\subset\mathcal X\times[0,\infty)\) be a bounded set of admissible inputs.

\begin{theorem}[Finite-depth input stability]
\label{stable-analysis}
Under the preceding assumptions, for each finite depth \(K\), there exist constants \(C_{\tau,K},C_{s,K}\geq0\) such that
\begin{equation}
\bigl\|
\mathcal G_\sigma(\tau,s_\sigma;\theta)
-
\mathcal G_\sigma(\tau',s_\sigma';\theta)
\bigr\|_{\mathcal U}
\leq
C_{\tau,K}\|\tau-\tau'\|_{\mathcal X}
+
C_{s,K}|s_\sigma-s_\sigma'|
\label{eq:input_stability}
\end{equation}
for all \((\tau,s_\sigma),(\tau',s_\sigma')\in\mathcal D\). The constants depend only on \(\mathcal D\), the discretization and the fixed network realization.
\end{theorem}

The proof is provided in Appendix~\ref{stable}.


\begin{remark}
Theorem~\ref{stable-analysis} concerns perturbation stability of the learned reconstruction map \cite{Hagemann_2021,combettes2020lipschitz}. The constants \(C_{\tau,K}\) and \(C_{s,K}\) quantify its sensitivity to perturbations in the transient measurement and noise descriptor, respectively, and  may increase with \(K\).
\end{remark}

\subsection{Recurrent Reconstruction and Convergence}
The reconstruction map \(\mathcal G_\sigma(\tau,s_\sigma;\theta)\)
uses a prescribed latent initialization to produce a volume in a single depth-\(K\) evaluation. We extend this construction by allowing the initial latent state to depend on a supplied reconstruction. Specifically, channel matching maps the current volume to the initial latent state, which is then processed
by the same trained latent hierarchy, anchoring map, and volume projection. This defines the state-update map
\[
\mathcal T_{\sigma,\theta}:
\mathcal U\times\mathcal D\rightarrow\mathcal U,
\]
where \(\mathcal G_\sigma\) provides the initial reconstruction, while \(\mathcal T_{\sigma,\theta}\) updates the reconstruction for fixed conditioning inputs. For a given \((\tau,s_\sigma)\in\mathcal D\), we consider
\begin{equation}
u^{(0)}
=
\mathcal G_\sigma(\tau,s_\sigma;\theta),
\qquad
u^{(n+1)}
=
\mathcal T_{\sigma,\theta}(u^{(n)};\tau,s_\sigma),
\quad n\geq0.
\label{eq:nano_fixed_point_iteration}
\end{equation}
The measurement-derived conditioning features, noise
embedding, LCT anchor, trained parameters, and SFE
weights remain fixed throughout the iteration.
Each outer step requires an additional depth-\(K\)
evaluation without introducing additional trainable
parameters. The procedure is summarized in Algorithm~\ref{alo-iter}. The recurrent scheme seeks a fixed point
\begin{equation}
u^\ast
=
\mathcal T_{\sigma,\theta}(u^\ast;\tau,s_\sigma),
\label{eq:nano_equilibrium_relation}
\end{equation}
which need not coincide with
\(\mathcal G_\sigma(\tau,s_\sigma;\theta)\).
The following result gives a sufficient condition
for convergence.
\begin{algorithm}[htbp]
\caption{Recurrent NANO reconstruction}
\label{alo-iter}
\begin{algorithmic}[1]
\Require Inputs \((\tau,s_\sigma)\in\mathcal D\),
trained parameters \(\theta\) fixed SFE weights
and an integer \(S\geq 1\)
\State \(u^{(0)}
\gets\mathcal G_\sigma(\tau,s_\sigma;\theta)\)
\For{\(n=1,\ldots,S\)}
\State \(u^{(n)}
\gets\mathcal T_{\sigma,\theta}
(u^{(n-1)};\tau,s_\sigma)\)
\EndFor
\State \Return \(u^{(S)}\)
\end{algorithmic}
\end{algorithm}

\begin{theorem}[Convergence of recurrent reconstruction]
\label{iter-convergence}
Let \((\tau,s_\sigma)\in\mathcal D\) be given, with
the trained parameters and SFE weights held fixed.
Assume that
\(\mathcal T_{\sigma,\theta}(\,\cdot\,;\tau,s_\sigma)\)
maps the Hilbert space \(\mathcal U\) into itself,
and that all conditioning features and the LCT anchor
are independent of the state.
Suppose that channel matching, the complete latent
hierarchy, anchoring, and subsequent volume projection
admit finite global Lipschitz bounds
\(C_{\mathbf l}\), \(L_{\Phi,K}\),
\(C_{\mathrm a}\) and \(C_{\mathbf p}\),
respectively, with respect to their state arguments.
If
\begin{equation}
q_K
:=
C_{\mathbf p}C_{\mathrm a}L_{\Phi,K}C_{\mathbf l}
<1,
\label{eq:contraction_factor}
\end{equation}
then the update has a unique fixed point
\(u^\ast\in\mathcal U\).
For every initial state \(u^{(0)}\in\mathcal U\),
the recurrence in \eqref{eq:nano_fixed_point_iteration}
converges to \(u^\ast\), with
\begin{equation}
\|u^{(n)}-u^\ast\|_{\mathcal U}
\leq q_K^n\|u^{(0)}-u^\ast\|_{\mathcal U},
\qquad n\geq1.
\label{eq:recurrent_convergence_rate}
\end{equation}
\end{theorem}

\begin{proof}
Let $x,y\in\mathcal U$, and consider two evaluations
with identical conditioning inputs.
The state-to-latent initialization mapping gives
\[
\|\widehat u^0-\widehat v^0\|
\leq C_{\mathbf l}\|x-y\|_{\mathcal U}.
\]

We apply the layerwise estimates underlying
Theorem~\ref{stable-analysis} with the conditioning
variables held fixed.
Omitting the scale index, the modulation operator
takes the form
\[
\mathcal S_{\sigma,k}(w,a;\theta)=q_k(a)\odot w.
\]
For fixed $a$, let $Q_k=\|q_k(a)\|_\infty$,
with broadcasting included.
Then
\[
\|\mathcal S_{\sigma,k}(w,a;\theta)
-\mathcal S_{\sigma,k}(z,a;\theta)\|
\leq Q_k\|w-z\|.
\]
Since the measurement-derived features are also fixed,
\eqref{fixed-point} yields
\[
\|\widehat u^{k+1}-\widehat v^{k+1}\|
\leq
\bigl(1+|\xi_k(\theta)|L_{\mathrm{sc},k}Q_k\bigr)
\|\widehat u^k-\widehat v^k\|,
\]
where $L_{\mathrm{sc},k}$ is a global Lipschitz
bound for the corresponding spectral block.
This estimate holds for arbitrary latent states.

Combining these estimates with the global bounds
for restriction, prolongation, skip connections,
and fixed SFE gives
\[
\|\widehat u^K-\widehat v^K\|
\leq
L_{\Phi,K}\|\widehat u^0-\widehat v^0\|
\leq
L_{\Phi,K}C_{\mathbf l}\|x-y\|_{\mathcal U}.
\]
Here, $L_{\Phi,K}$ accounts for the complete
latent hierarchy.

The LCT anchor is independent of the state and
therefore cancels in differences.
With $\mu(\theta)\in[0,1]$ fixed, the anchoring map
admits the bound
\[
C_{\mathrm a}
=
\bigl(1-\mu(\theta)\bigr)
\operatorname{Lip}
\bigl(\operatorname{Conv}_{\mathbb C}(\,\cdot\,;\theta)\bigr).
\]
Applying the subsequent volume-projection bound yields
\begin{equation}
\bigl\|
\mathcal T_{\sigma,\theta}(x;\tau,s_\sigma)
-
\mathcal T_{\sigma,\theta}(y;\tau,s_\sigma)
\bigr\|_{\mathcal U}
\leq
C_{\mathbf p}C_{\mathrm a}L_{\Phi,K}C_{\mathbf l}
\|x-y\|_{\mathcal U}
=
q_K\|x-y\|_{\mathcal U}.
\label{eq:recurrent_lipschitz}
\end{equation}
Under \eqref{eq:contraction_factor}, the update
is a contraction on the complete space $\mathcal U$.
The Banach fixed-point theorem \cite{smart1980fixed}
therefore gives a unique fixed point $u^\ast\in\mathcal U$.
Applying \eqref{eq:recurrent_lipschitz} successively
with one argument equal to $u^\ast$ proves
\eqref{eq:recurrent_convergence_rate}.
\end{proof}

\begin{remark}
Input stability and recurrent convergence use the same
layerwise estimates but concern input and state
perturbations, respectively.
The contraction condition $q_K<1$ does not follow
from input stability, it is an additional sufficient
condition for convergence to a unique fixed point.
\end{remark}

\section{Numerical experiments}\label{numerical-experiment}
In this section, we evaluate NANO on simulated and real-world NLOS measurements. The experiments assess reconstruction accuracy under multiple synthetic noise levels, robustness to changes in real-data exposure time and spatial sampling density, the contributions of the main architectural components, computational cost, and empirical finite-depth sensitivity. We additionally examine the finite-iteration
behaviour of the recurrent formulation. These experiments characterize the performance of the proposed noise-conditioned reconstruction framework under the evaluated acquisition settings.

\subsection{Experimental setup}
We implement NANO in PyTorch and train it for 30 epochs
with a batch size of 1 using AdamW, an initial learning
rate of $10^{-4}$, and exponential learning-rate decay.
The reconstruction network uses five scales with two
refinement steps per scale.
We use the synthetic dataset from \cite{chen2020learned},
comprising 2,712 training samples and 291 test samples
generated from 277 motorcycle models in ShapeNet
\cite{chang2015shapenet}.
Each transient is represented on a $128\times128$
scan grid with 512 temporal bins of width 33 ps.

We augment the training transients with additive and
multiplicative Gaussian perturbations, introducing
signal-independent noise and intensity-dependent
fluctuations, respectively.
A constant background offset of $0.03$ is also included.
This construction provides empirical noise augmentation
rather than a direct simulation of Poisson photon counts.
The pretrained NLE module estimates the noise scale
$\sigma$ from each corrupted transient.
The conditioning descriptor $s_\sigma$ is then computed
using \eqref{eq:noiselevel} and supplied to
$\mathcal G_\sigma$.

The loss weights are set to $\lambda_{\mathrm d}=0$
and $\lambda_{\mathrm{tv}}=10^{-7}$ for the first
12 epochs, and to $\lambda_{\mathrm d}=0.5$ and
$\lambda_{\mathrm{tv}}=10^{-8}$ for the remaining epochs.
All experiments, including retraining the publicly
available LFE, NLOST, and LPP models, are conducted
on a workstation equipped with four NVIDIA RTX A6000 GPUs.

\paragraph{Datasets}
We evaluate the proposed method on three benchmark datasets, including the Rendering dataset \cite{chen2020learned}, the Stanford dataset \cite{lindell2019wave}, and the USTC dataset \cite{li2023nlost}. The
Rendering dataset contains synthetic transient measurements over a $2\,\mathrm{m}\times2\,\mathrm{m}$ relay wall with 512 temporal bins of 33 ps. The Stanford dataset provides real photon-counting measurements acquired over a $2\,\mathrm{m}\times2\,\mathrm{m}$ relay wall with a temporal resolution of 32 ps. Importantly, several scenes are captured at multiple exposure times, providing measurements with different photon counts and noise levels. For preprocessing, the Stanford measurements are first aligned using the provided time-of-flight calibration, cropped to 512 temporal bins, and spatially downsampled to $128\times128$. The USTC dataset contains real-world transient measurements with a spatial resolution of $128\times128$ and 512 temporal bins of 32 ps. For consistency, all datasets are finally represented at a common resolution of $128\times128\times512$ before being fed into the reconstruction network.

The reconstruction performance is evaluated from both intensity and depth perspectives. Peak signal-to-noise ratio (PSNR) and structural similarity index (SSIM) are used to assess the reconstructed intensity images, while root mean square error (RMSE) and mean absolute distance (MAD) are adopted for depth evaluation.

\begin{figure}[ht]
\begin{center}			 
\includegraphics[width=0.8\linewidth]{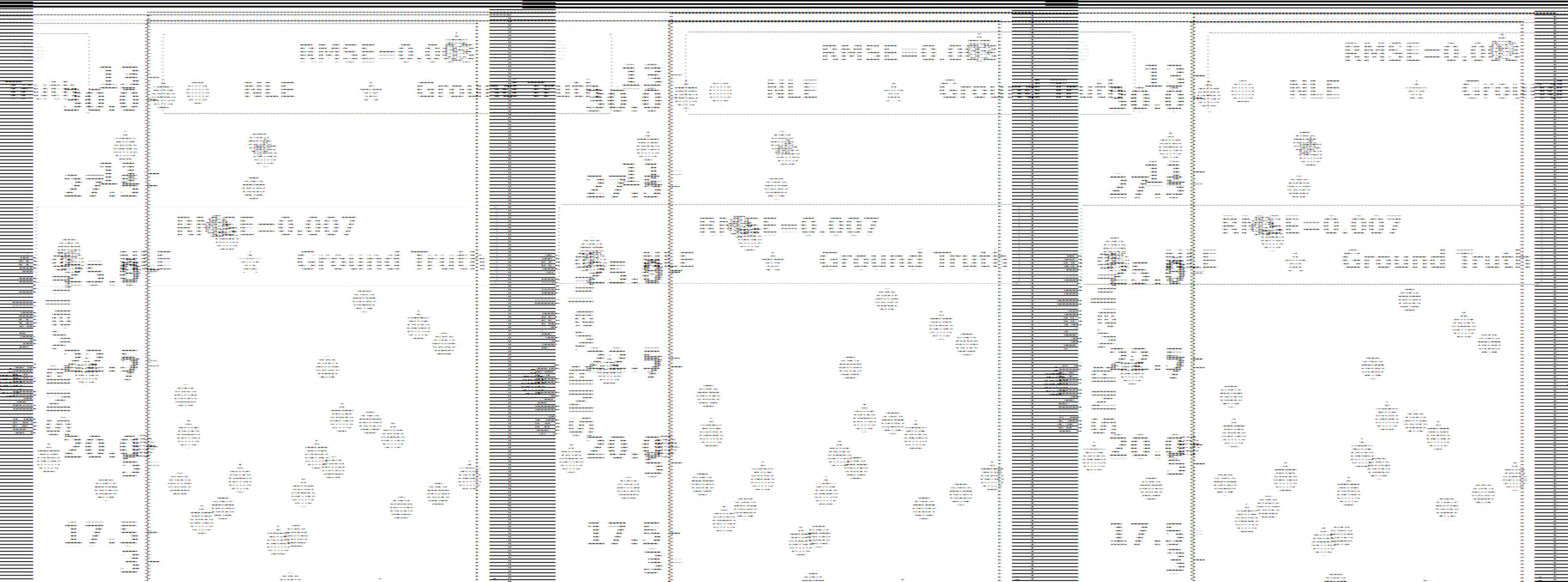}
	\end{center}
	\caption{Evaluation of the noise level estimation module on simulated data. The left shows the noise range consistent with the training dataset, while the right illustrates a higher noise level.}
	\label{fig:NLE}
\end{figure}

\paragraph{Comparison methods}
To demonstrate the effectiveness of NANO, we conduct a comprehensive evaluation against six state-of-the-art baselines on both synthetic and real-world datasets. These include three representative physics-based classical methods: LCT \cite{o2018confocal}, FK \cite{lindell2019wave} and PF \cite{liu2019non}, and three learning-based methods: LFE \cite{chen2020learned}, NLOST \cite{li2023nlost} and LPP \cite{sun2025generalizable}. For synthetic evaluations, each learning-based model is assessed under both original fixed-noise (FN) and our varying-noise (VN) configurations. For real-world datasets, all learning-based models are naturally evaluated under the VN protocol to ensure a fair and unified comparison under real noise perturbations.

\subsection{Results on simulated data}
For the corrupted measurement, we compute the observation-to-noise ratio (ONR) via
\begin{equation}
\mathrm{ONR}=-10\log_{10}\left(s_\sigma\right),
\label{eq:experimental_snr}
\end{equation}
where $s_\sigma$ is denoted by \eqref{eq:noiselevel}.
Figure~\ref{fig:NLE} evaluates the accuracy of the noise-level estimator by comparing the ONR values derived from the estimated noise scale with the prescribed reference values. Across the evaluated noise regimes, the estimated values remain close to their references, including the ONR range outside that used for pretraining, as indicated by the reported RMSE values. These results support the use of the estimated noise descriptor as a conditioning variable for selecting the corresponding regularization regime. The influence of this conditioning on the reconstructed solution is evaluated in the subsequent experiments.

We generate corrupted measurements from a clean
transient $\tau_0$ according to
\[
\tau_g=\tau_0+b+c_1\varepsilon_1,
\qquad
\tau_{\mathrm{noisy}}
=
\tau_g+c_2(\tau_g\odot\varepsilon_2),
\]
where $\varepsilon_1$ and $\varepsilon_2$ are
independent standard Gaussian noise fields,
$b=0.03$ is a constant background offset, and
$c_1,c_2\geq0$ control the additive and multiplicative
perturbations, respectively.   We consider two corruption settings, $c_1=c_2=0.1$ and $c_1=c_2=0.2$, yielding the mean ONRs of approximately $20$ dB and $15$ dB, respectively. Since signal power and estimated noise vary per sample, all reported ONR values represent test-set averages.

\begin{figure}[t]
    \begin{center}			  \includegraphics[width=1.0\linewidth]{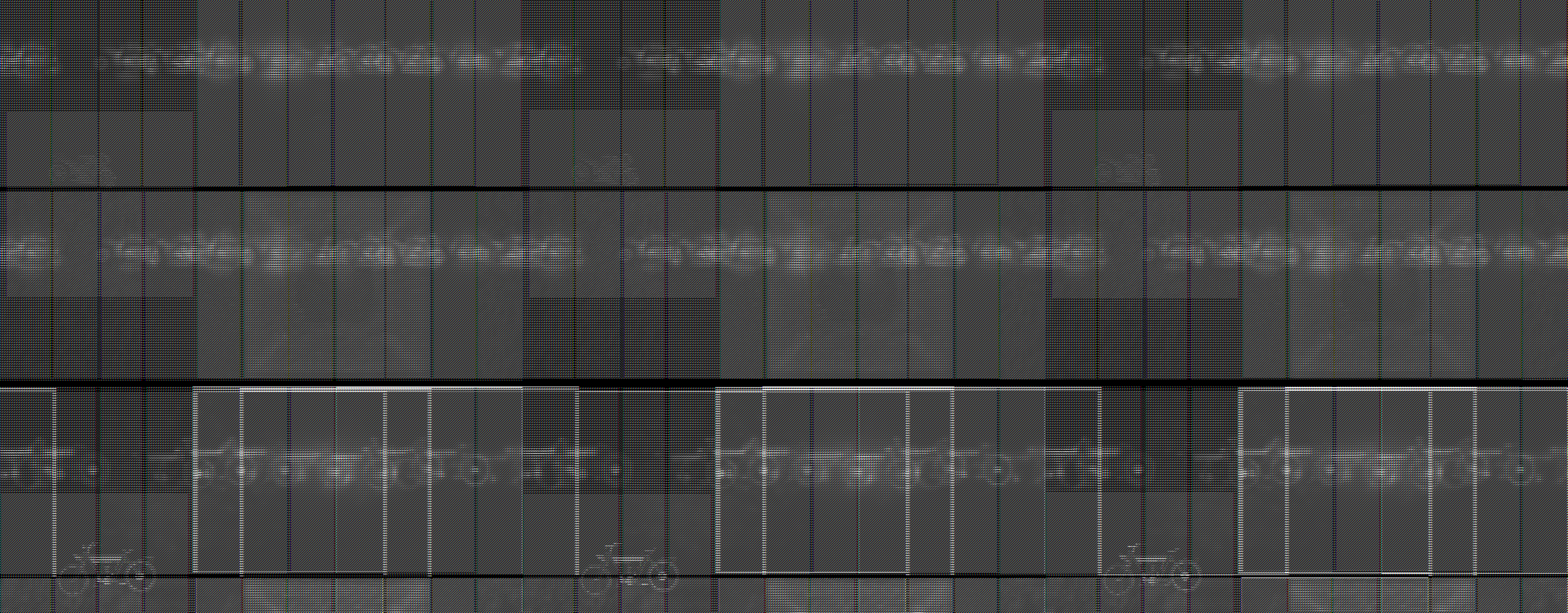}
	\end{center}
    \vspace{-.2cm}
	\caption{Reconstruction results on the test set of the simulated dataset under different noise conditions. The first and third rows show the results under a high-quality measurement condition with $\mathrm{ONR}=20$ dB, while the second and fourth rows correspond to the degraded measurements with $\mathrm{ONR}=15$ dB. For each learning-based baseline, FN denotes the model trained using its original fixed-noise setting, and VN denotes the model retrained using the varying-noise setting adopted in our experiments.}
	\label{noise-128}
\end{figure}

Tables~\ref{table:time} and~\ref{table:noise} summarise
the quantitative results, with representative
reconstructions shown in Figure~\ref{noise-128}.
Under the higher estimated-ONR condition, PF attains
the highest albedo PSNR among the analytical methods,
while FK gives the lowest depth RMSE.
Among the learned baselines, LPP-FN provides the highest
albedo PSNR and SSIM.
NANO improves PSNR by more than 4 dB relative to LCT
and FK, but only marginally relative to LPP-FN
(0.05 dB), while also achieving higher SSIM.
For depth reconstruction, NANO yields the lowest MAD,
whereas NLOST-VN achieves the lowest RMSE.
These results indicate improved albedo reconstruction
and competitive  depth accuracy.

\begin{table}[t]
\footnotesize
\setlength{\tabcolsep}{2.8pt} 
\renewcommand{\arraystretch}{1.2} 
\caption{Quantitative comparison of reconstruction methods for intensity images and depth maps on simulated dataset ($\mathrm{ONR}=20\,\mathrm{dB}$), alongside backbone networks, memory consumption, and reconstruction runtime. }
\label{table:time}
\centering
\begin{tabular}{c|c c c|c c|c c}
\toprule
\multirow{2}{*}{\textbf{Methods}} & \multirow{2}{*}{\textbf{ Backbone}} & \multirow{2}{*}{\textbf{Memory}} & \multirow{2}{*}{\textbf{Time}} & \multicolumn{2}{c|}{\textbf{Albedo}} &
\multicolumn{2}{c}{\textbf{Depth}}\\
\cline{5-8}
 &&&& \textbf{PSNR$\uparrow$} &\textbf{SSIM$\uparrow$} & \textbf{RMSE$\downarrow$} &\textbf{MAD$\downarrow$}  \\
\midrule
LCT \cite{o2018confocal} & Physics  & 6GB  &0.05s & 23.13 & 0.72  &0.619 &0.610 \\
FK \cite{lindell2019wave} & Physics  & 7GB  &0.06s  &23.12  &0.75 &0.522 &0.501 \\
PF \cite{liu2019non} & Physics  & 10GB  & 0.12s &24.41& 0.74 &0.701 &0.653\\
LFE-FN \cite{chen2020learned} & CNN  & 4GB  & 0.03s  &27.07  &0.83 & 0.077    &0.047 \\
LFE-VN  & CNN  & 4GB  & 0.03s  &26.14  &0.86 & 0.073    &0.025 \\
NLOST-FN \cite{li2023nlost} & Transformer  & 38GB  &0.59s   &27.84  &0.87  & 0.068 & 0.025 \\
NLOST-VN  & Transformer  & 38GB  &0.59s   &27.51  &0.71  &  \textbf{0.044} & 0.024 \\
LPP-FN \cite{sun2025generalizable} & CNN  & 17GB  &0.35s   &\underline{27.99}  &\underline{0.89}  & 0.060 & 0.019 \\
LPP-VN  & CNN  & 17GB  &0.35s   &26.57  &0.77  & 0.055 & \underline{0.015} \\
\midrule
 NANO & CNN & 14GB &  0.53s & \textbf{28.04} &   \textbf{0.92}  & \underline{0.054} &  \textbf{0.013} \\
\bottomrule
\end{tabular}
\end{table}
\begin{table}[ht]
\centering
\footnotesize
\setlength{\tabcolsep}{2.2pt}  
\renewcommand{\arraystretch}{1.3} 
\caption{Quantitative comparison of different methods for intensity image reconstruction on the test set ($\mathrm{ONR}=15\,\mathrm{dB}$).}
\label{table:noise}
\begin{tabular}{c|c cc cccc c c|c}
\toprule
\textbf{} & \textbf{LCT}  & \textbf{FK} & \textbf{PF}  & \textbf{LFE-FN}& \textbf{LFE-VN} & \textbf{NLOST-FN} & \textbf{NLOST-VN}& \textbf{LPP-FN}& \textbf{LPP-VN} &   \textbf{NANO} \\
\midrule
\textbf{PSNR $\uparrow$} & 23.12 & 21.09 & 23.12 & 24.06 & 26.49 & 25.35 & 22.16 & 24.75 & 26.20  &   \textbf{28.01} \\
\textbf{SSIM $\uparrow$} & 0.73  & 0.64  & 0.70  & 0.57  & 0.89 & 0.53  & 0.75  & 0.67  & 0.79 &   \textbf{0.92} \\
\bottomrule
\end{tabular}
\end{table}

When the ONR decreases to 15 dB, the increased perturbations makes the inverse reconstruction substantially more challenging. For learning-based approaches, retraining with our noise augmentation improves robustness under degraded measurements, however, this improvement comes at the cost of reduced reconstruction fidelity under high-quality conditions. As illustrated in the second scene of Figure \ref{noise-128}, the noise-augmented models exhibit excessive smoothing and loss of fine structures, as the networks learn a more conservative inverse mapping to accommodate diverse noisy observations. In contrast, NANO maintains consistent reconstruction quality across both ONR conditions by explicitly conditioning the inverse operator on the estimated noise descriptor.

\begin{figure}[t]
      \begin{center}			
\includegraphics[width=0.8\linewidth]{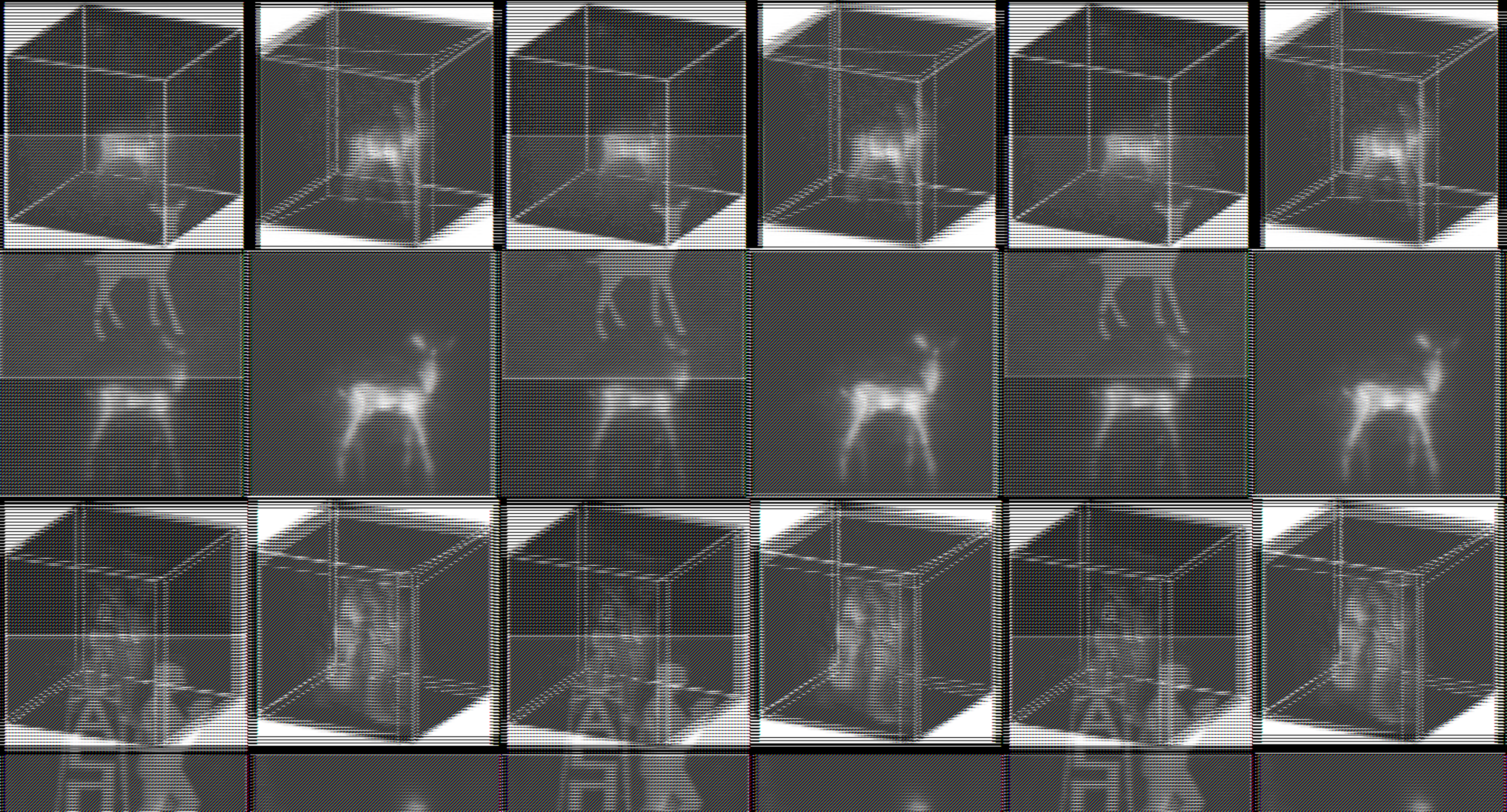}
	  \end{center}
      \vspace{-0.2cm}
	\caption{Visualization comparisons on hidden scenes from the public real-world data \cite{li2023nlost}, where both maximum projection and 3D reconstruction are provided for different methods.}
	\label{manletter-128}
\end{figure}
\begin{figure}[ht]
    \centering
    \includegraphics[width=0.95\linewidth]{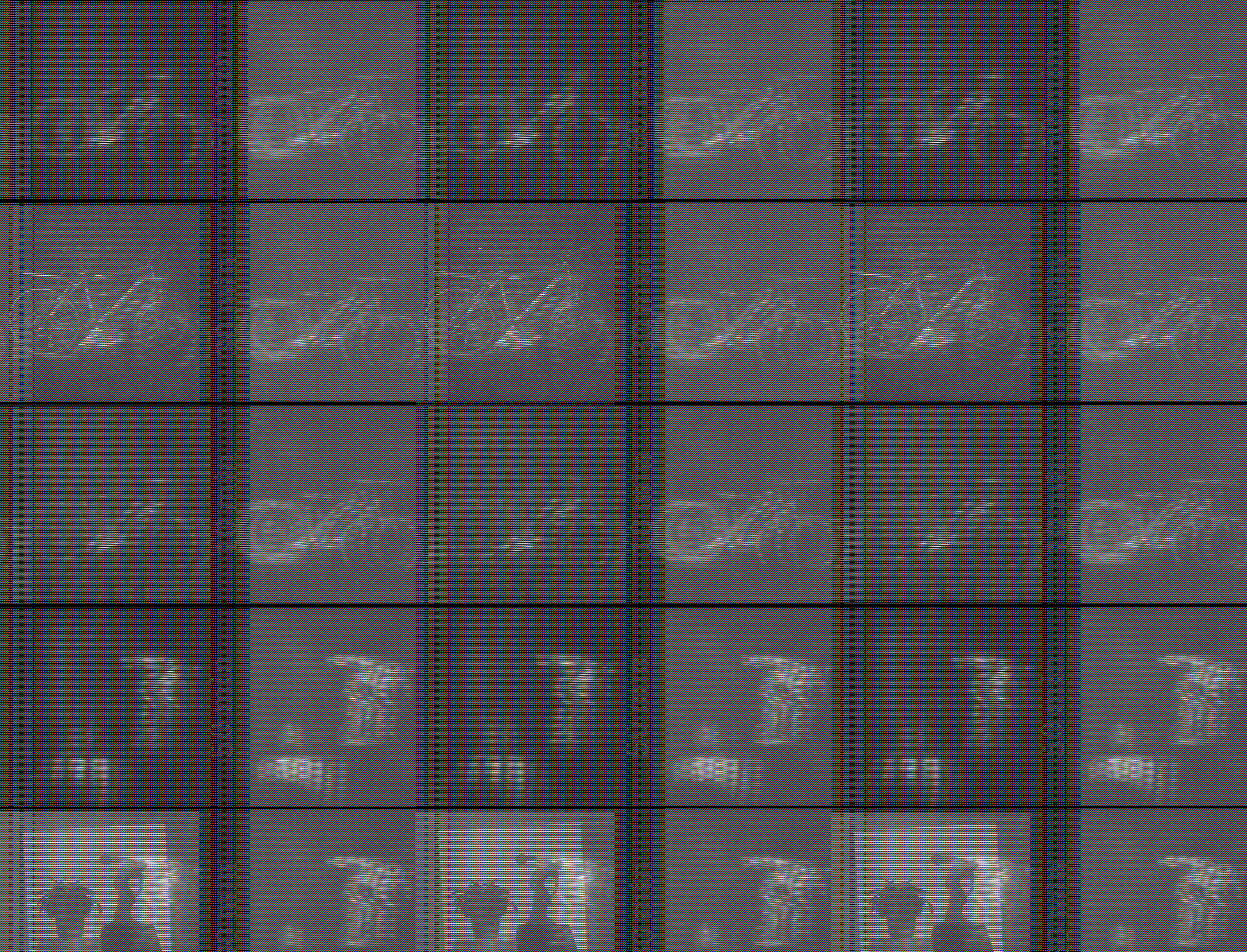}
    \vspace{-0.2cm}
    \caption{Comparison of reconstruction results under different exposure times on the real-world dataset provided by \cite{lindell2019wave}.}
    \label{fig:realnoise}
\end{figure}

\subsection{Results on real-world data}
The qualitative results of the real data and  the corresponding 3D reconstructions are illustrated in Figure \ref{manletter-128}. Our method effectively recovers fine spatial structures and sharp boundaries in hidden scenes, particularly visible in the deer's legs and textual regions. While LCT utilizes confocal time-of-flight relationships for hidden target localization, FK and PF incorporate wave physics to improve resolution. Nevertheless, their analytical formulations remain vulnerable to real-world noise and attenuation, yielding degraded boundaries and incomplete structures. Although LFE offers slight gains, it fails to recover sharp details. NLOST and LPP improve resolution using spatio-temporal and phasor-field priors, but remain sensitive to noise perturbations.
In comparison, NANO combines noise-dependent feature
modulation with an LCT-derived reconstruction anchor.
The results show more coherent structures under the
evaluated acquisition conditions, using the same
trained model.

Figure \ref{fig:realnoise} presents the 2D albedo reconstructions from real-world measurements acquired with different exposure times, evaluating the robustness of our method under varying photon-count conditions. Reduced exposure severely limits detected photon counts, thereby increasing measurement fluctuations and heightening inversion instability. Physics-based methods exhibit amplified noise artifacts and degraded structural recovery. Without retraining, other learning-based methods also experience performance degradation, while our method still delivers visually clearer structures.

\subsection{Results on down-sampled data}
\begin{figure}[t]
      \begin{center}     \includegraphics[width=0.9\linewidth]{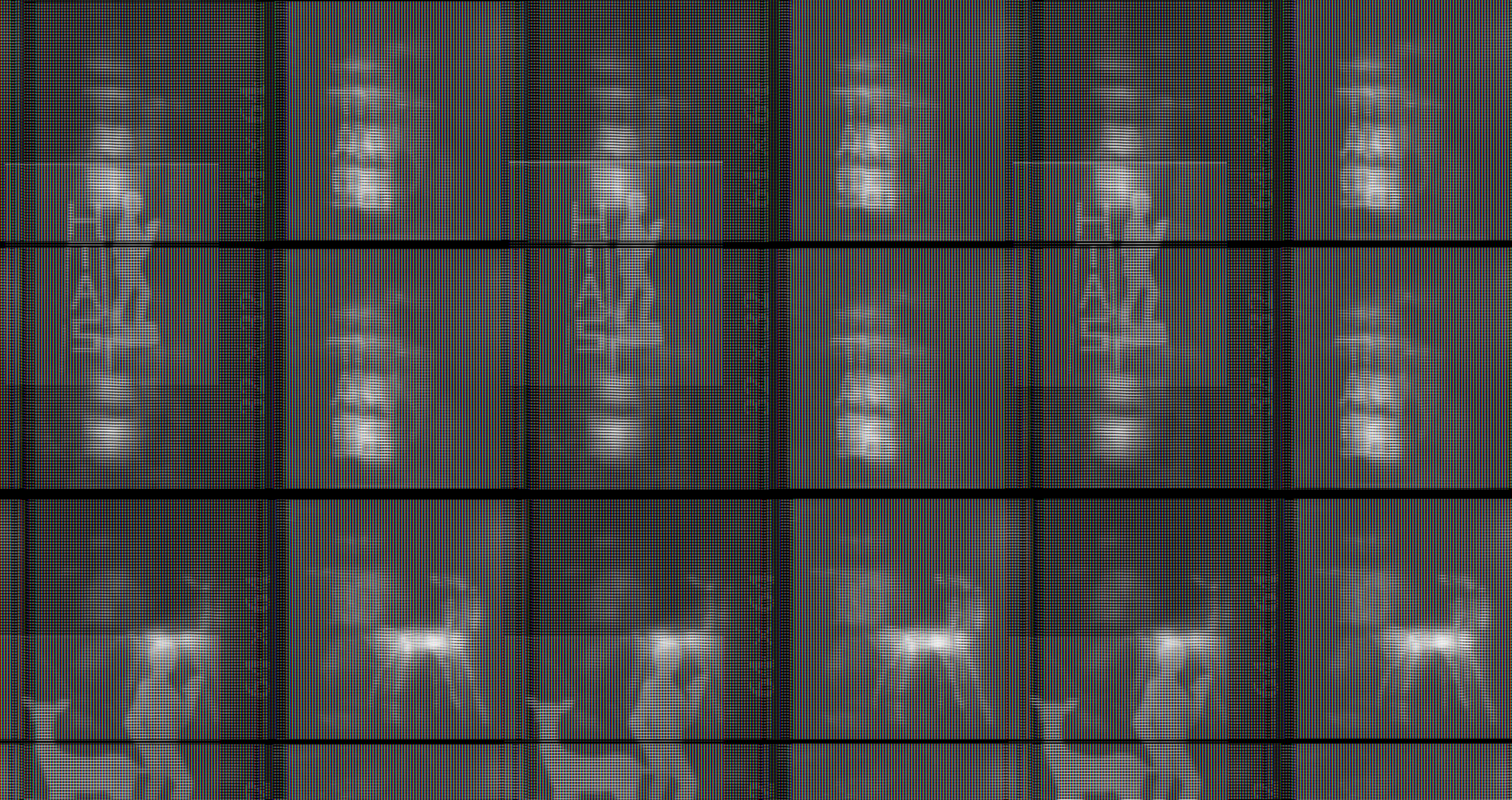}
	  \end{center}
      \vspace{-0.2cm}
	\caption{Reconstructed intensity images of real-world scenes \cite{li2023nlost} under different spatial sampling densities.}
	\label{downsample-ustc}
\end{figure}

\begin{figure}[ht]
      \begin{center} \includegraphics[width=0.9\linewidth]{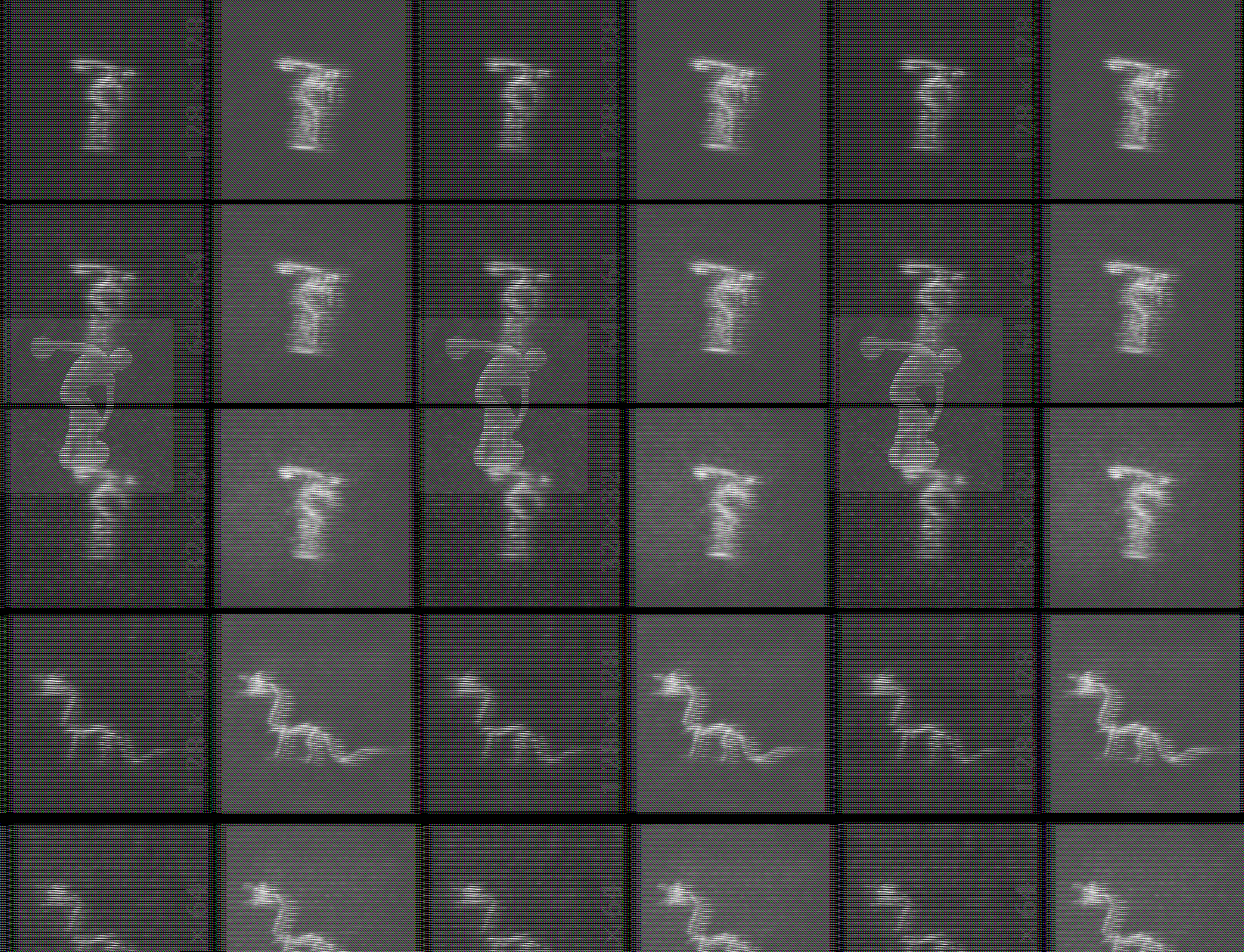}
	  \end{center}
      \vspace{-0.2cm}
	\caption{Reconstructed intensity images of real-world scenes acquired with a 10-minute exposure \cite{lindell2019wave} under different spatial sampling densities obtained by subsampling the measured transients.}
	\label{downsample-bike}
\end{figure}

We further evaluate the robustness of NANO under reduced spatial sampling densities using low-exposure measurements. Specifically, the spatial resolution of the transient measurements is uniformly reduced from the original
$128\times128$ grid to $64\times64$ and $32\times32$, and then interpolated to $128\times128$ using trilinear interpolation to match the NLE network input. Figures \ref{downsample-ustc} and \ref{downsample-bike} present the corresponding reconstruction results under different sampling densities. As the spatial
sampling density decreases, the available transient information becomes more
limited. LFE suffers from degraded intensity recovery, while NLOST and LPP exhibit more noise artifacts under sparse measurements. In contrast, NANO achieves more consistent
structural recovery by effectively exploiting the available transient information through neural operator learning. 

\begin{table}[t]
\centering
\caption{
Ablation results on simulated data. Each result is reported as H/L,
where H and L correspond to the $\mathrm{ONR}=20$ dB and
$\mathrm{ONR}=15$ dB, respectively.
LM: lifting module; MLP-TL: MLP-based transient lifting;
GL-TL: global-local transient lifting;
NLE: noise-level estimation;
SFE: stochastic frequency encoding;
PA: physics anchoring.
}
\label{table:ablation}
\footnotesize
\setlength{\tabcolsep}{2.5pt}
\renewcommand{\arraystretch}{1.10}
\begin{adjustbox}{max width=\linewidth}
\begin{tabular}{@{}c|cc|ccc|cc|cc@{}}
\toprule
& \multicolumn{2}{c|}{\textbf{LM}}
& \multicolumn{3}{c|}{\textbf{Components}}
& \multicolumn{2}{c|}{\textbf{Albedo}}
& \multicolumn{2}{c}{\textbf{Depth}}
\\
\cmidrule(lr){2-3}\cmidrule(lr){4-6}\cmidrule(lr){7-8}\cmidrule(l){9-10}
\textbf{Case}
& \textbf{MLP-TL}& \textbf{GL-TL}& \textbf{NLE}& \textbf{SFE}& \textbf{PA}& \makecell{\textbf{PSNR} $\uparrow$\\[-1pt]\textbf{(H/L)}}& \makecell{\textbf{SSIM} $\uparrow$\\[-1pt]\textbf{(H/L)}}& \makecell{\textbf{RMSE} $\downarrow$\\[-1pt]\textbf{(H/L)}}& \makecell{\textbf{MAD} $\downarrow$\\[-1pt]\textbf{(H/L)}}
\\
\midrule
\textbf{I}& \ding{55}& \ding{55}& \ding{55}& \ding{51}& \ding{55}& 25.34 / 23.49 & 0.87 / 0.84 & 0.126 / 0.176 & 0.031 / 0.053
\\
\textbf{II}& \ding{51}& \ding{55}& \ding{55}& \ding{51}& \ding{55}& 24.68 / 24.33& 0.85 / 0.81& 0.103 / 0.125& 0.053 / 0.050
\\
\textbf{III}& \ding{55}& \ding{51}& \ding{55}& \ding{51}& \ding{55}& 26.21 / 25.71& 0.87 / 0.82& 0.082 / 0.085& 0.030 / 0.031
\\
\midrule
\textbf{IV}& \ding{55}& \ding{51}& \ding{51}& \ding{55}& \ding{55}& 26.48 / 26.54& 0.87 / 0.84& 0.070 / 0.073& 0.025 / 0.030\\
\textbf{V}& \ding{55}& \ding{51}& \ding{51}& \ding{51}& \ding{55}& 28.01 / \textbf{28.03}& 0.90 / 0.88& 0.065 / 0.065& 0.021 / 0.023\\
\midrule
\textbf{VI}& \ding{55}& \ding{51}& \ding{51}& \ding{51}& \ding{51}& \textbf{28.04} / 28.01& \textbf{0.92 / 0.92}& \textbf{0.055 / 0.053}& \textbf{0.014 / 0.014}
\\
\bottomrule
\end{tabular}
\end{adjustbox}
\end{table}

\subsection{Ablation studies}

\begin{figure}[ht]
      \begin{center}			
\includegraphics[width=0.8\linewidth]{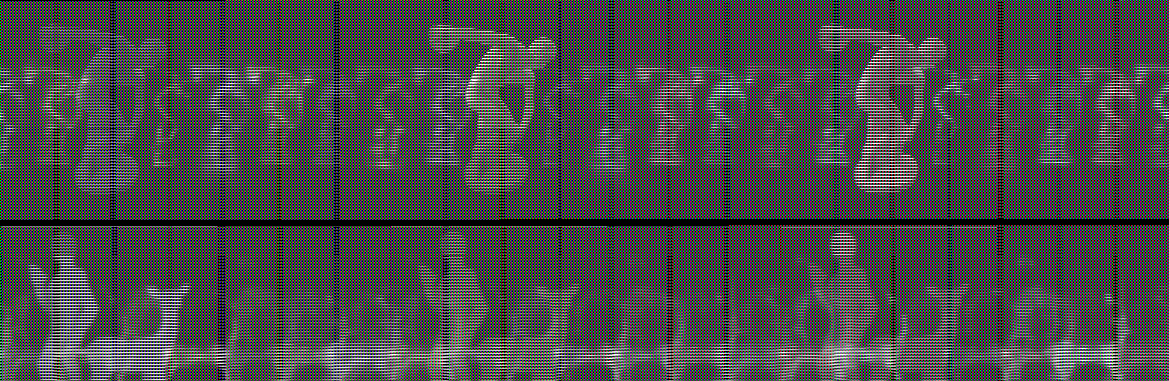}
	  \end{center}
      \vspace{-0.2cm}
	\caption{Qualitative ablation results on real-world measurements. From left to right: reference scene, reconstruction without the local transient-lifting branch, without the global transient-lifting branch, without stochastic frequency encoding, without noise-level estimation, without physics anchoring, and the complete NANO model.}
	\label{ablation}
\end{figure}

We evaluate the contributions of the reconstruction
architecture and noise conditioning through the ablations
in Table~\ref{table:ablation}.
Compared with direct reconstruction in Case I,
the latent-space formulation in Case II reduces the
difference in mean PSNR between the two tested noise
ranges from 1.85 dB to 0.35 dB.
This indicates reduced performance variation across
these ranges, although the mean PSNR in the broader
range decreases.
Replacing the fixed conditioning value in Case III
with the measurement-specific NLE descriptor in Case V
improves the mean PSNR in both ranges and reduces
the difference between them.
The real-data comparisons in Figure~\ref{ablation}
also show improved structural recovery with NLE
conditioning.

Compared with MLP-based lifting in Case II,
GL-TL in Case III improves both albedo and depth
metrics.
The global branch aggregates spatiotemporal information
across relay-wall locations, while the local branch
processes temporal signatures within spatial
neighbourhoods.
Table~\ref{table:glablation} shows that combining
the branches improves albedo PSNR, SSIM, and depth
RMSE relative to either branch alone, although
depth MAD does not improve.
Adding SFE in Case V improves the albedo and depth
metrics relative to Case IV.
Finally, the LCT-derived anchor provides complementary
model-based reconstruction information.
Its inclusion in Case VI reduces both depth RMSE
and MAD relative to Case V.

Figure~\ref{ablation} further presents qualitative ablations on real-world measurements by removing one component at a time from the complete NANO model. Removing either the global or local transient-lifting branch results in less complete structural recovery and more pronounced artifacts. The variants without NLE or SFE show reduced structural clarity, while removing the physics anchor reduces geometric continuity. The complete model yields more coherent reconstructions in the evaluated real-world scenes.
\begin{table}[ht]
\footnotesize
\caption{Ablation results on simulated data under the higher-ONR setting. G-TL and L-TL: global and local spatiotemporal transient lifting. }
\label{table:glablation}
\centering
\begin{tabular}{c|c c|c c|c c} 
\toprule
\multirow{2}{*}{\textbf{  }} & \multirow{2}{*}{\textbf{G-TL}} & \multirow{2}{*}{\textbf{L-TL}} & \multicolumn{2}{c|}{\textbf{Albedo}}& \multicolumn{2}{c}{\textbf{Depth}}  \\
\cline{4-7}
 & &&\textbf{PSNR$\uparrow$} & \textbf{SSIM$\uparrow$} & \textbf{RMSE$\downarrow$} &\textbf{MAD$\downarrow$} \\
\midrule
\textbf{III-I}  & \ding{51}   & \ding{55}  & 25.66 & 0.86 &0.091 &\textbf{0.028} \\
\textbf{III-II} & \ding{55} & \ding{51} &25.93 & 0.86 &0.085 &0.029 \\
\midrule
\textbf{III}    & \ding{51} & \ding{51}&\textbf{26.21} & \textbf{0.87}  &\textbf{0.082} & 0.030 \\
\bottomrule
\end{tabular}
\end{table}

\begin{table}[ht]
\footnotesize
\setlength{\tabcolsep}{4pt}  
\caption{Comparisons of different deep models about their training time, inference time, training memory and total parameters.}
\label{table:memory}
\centering
\begin{tabular}{c| c  c c  c  c  }
\toprule
\textbf{} & \textbf{LFE}  & \textbf{NLOST} & \textbf{LPP} &\textbf{NoSFE} &\textbf{NANO}\\
\midrule
\textbf{Training / Test } & 10h / 0.03s & 38h / 0.59s & 16h / 0.35s & 30h / 0.50s & 32h / 0.53s  \\
\textbf{Memory / Params} & 4G / 0.03M & 38G / 1M & 17G / 13M & 10G / 4M & 14G / 70M  \\
\bottomrule
\end{tabular}
\end{table}

\subsection{Computational efficiency and model complexity}

Table~\ref{table:memory} compares the computational cost of the learning-based reconstruction methods under the same hardware environment. LFE \cite{chen2020learned} provides the lowest computational cost, whereas NLOST \cite{li2023nlost} requires the longest training time and the largest memory allocation. NANO requires less inference memory than NLOST and LPP \cite{sun2025generalizable} and has an inference time comparable to NLOST, although its parameter count is larger.
The comparison between NoSFE and NANO shows that the additional parameters arise primarily from the MLP used in the stochastic frequency encoding. Since this encoding is applied at the coarsest latent scale, increasing the parameter count results in relatively moderate increases in inference time and memory consumption. Together with the ablation results in Table~\ref{table:ablation}, this comparison characterizes the computational trade-off introduced by SFE: it improves the reconstruction of spatial details while increasing the model size, without a proportional increase in inference cost.

\subsection{Empirical sensitivity and recurrent behaviour}

Theorem~\ref{stable-analysis} establishes finite-depth
input stability, while Theorem~\ref{iter-convergence}
gives a sufficient condition for recurrent convergence.
We examine the local sensitivity of the latent updates
and the finite-iteration behaviour of the recurrent
reconstruction on the evaluated samples. For each update stage \(k\), we compute the directional
perturbation ratio
\[
r_{\sigma,k}(w,\delta)
=
\frac{
\|\mathcal A_{\sigma,k}(w+\delta)
-\mathcal A_{\sigma,k}(w)\|_2
}{
\|\delta\|_2
},
\qquad \delta\neq0,
\]
where \(w\) is a latent state and \(\delta\) is a small
random perturbation.
The conditioning features and trained parameters are
held fixed and omitted from the notation.

\begin{figure}[t]
    \begin{center}			 
    \includegraphics[width=1\linewidth]{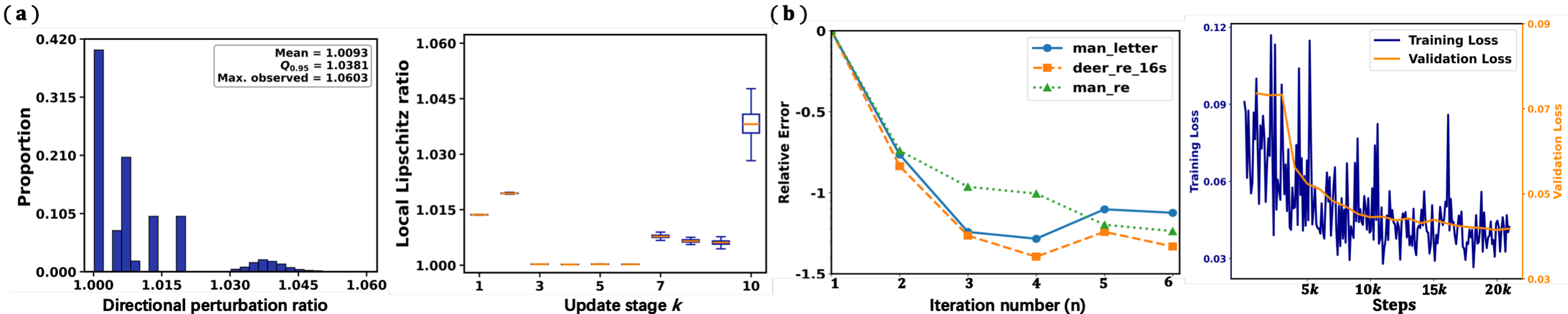}
	\end{center}
    \vspace{-0.2cm}
\caption{Empirical sensitivity and recurrent behaviour. (a) Pooled and stage-wise distributions of the empirical directional perturbation ratios for the latent updates on the simulated dataset. (b) Logarithmic relative residuals on the real-world dataset \cite{li2023nlost}, obtained by repeatedly applying a single trained iterator with fixed model parameters and conditioning inputs, together with the corresponding training and validation loss curves.}
\label{fig:convergence}
\end{figure}

Figure~\ref{fig:convergence}(a) reports the pooled
and stage-wise distributions of these ratios.
Across the evaluated states and perturbations,
the mean is \(1.0093\), the empirical \(0.95\)-quantile
is \(1.0381\), and the largest observed value is
\(1.0603\).
The ratios remain narrowly concentrated around unity across the update stages, indicating that small perturbations undergo only limited amplification through the latent refinement process. This result provides empirical evidence of controlled finite-depth sensitivity over the evaluated latent states and perturbation directions.

\begin{table}[ht]
\footnotesize
\setlength{\tabcolsep}{3pt} 
    \caption{Quantitative comparison of independently trained models with different reconstruction depths \(S\).}\label{table:convergence}
    \centering
    \begin{tabular}{c|c c c c c c}
        \toprule
        & \textbf{$\mathbf{S=1}$} & \textbf{$\mathbf{S=2}$} & \textbf{$\mathbf{S=3}$} & \textbf{$\mathbf{S=4}$} & \textbf{$\mathbf{S=5}$} & { \textbf{$\mathbf{S=6}$}} \\
        & \scriptsize (14G) & \scriptsize (18G) & \scriptsize (23G) & \scriptsize (26G) & \scriptsize (30G) & \textbf{\scriptsize (34G)} \\
        \midrule
        \textbf{PSNR $\uparrow$} & 27.98 & 28.26 & 28.21 & 28.21 & 28.22 & \textbf{28.36} \\
        \textbf{SSIM $\uparrow$} & 0.881  & 0.886  & 0.885  & 0.902  & 0.905  & { \textbf{0.910}}  \\
        \bottomrule
    \end{tabular}
\end{table}

We assess the effect of the recurrent depth $S$ by
training a separate model for each value of $S$ for
10 epochs, with all other training settings unchanged.
The recurrent updates follow Algorithm~\ref{alo-iter}.
As shown in Table~\ref{table:convergence}, larger
recurrent depths generally improve reconstruction
quality, although the gains are not monotonic.
These results compare independently trained
configurations rather than additional inference
iterations of a single fixed model.
Parameter sharing preserves the parameter count,
while the measured memory consumption increases
from 14 GB at $S=1$ to 34 GB at $S=6$.

Separately, we examine recurrent evaluations of a
fixed trained model, keeping the parameters \(\theta\),
conditioning inputs \((\tau,s_\sigma)\), and SFE
weights unchanged.
We record the logarithmic relative update
\[
r_n
=
\log_{10}
\left(
\frac{
\|u^{(n+1)}-u^{(n)}\|_{\mathcal U}
}{
\|u^{(n+1)}\|_{\mathcal U}
}
\right),
\]
for iterates with nonzero numerator and denominator.
Figure~\ref{fig:convergence}(b) shows a pronounced reduction in the relative residual during the initial recurrent evaluations, followed by small non-monotonic variations at lower levels. The residuals remain below their initial values throughout the evaluated iterations, indicating progressively smaller changes between successive reconstructions and controlled finite-iteration behaviour of the learned recurrent operator. These experiments characterize directional sensitivity
and finite-iteration behaviour on the evaluated samples;
they do not verify the global contraction condition
in Theorem~\ref{iter-convergence}.

\section{Conclusions}
\label{conclusions}
In this work, we have presented NANO, a noise-adapted neural operator for robust NLOS reconstruction. By conditioning the learned inverse operator on measurement-specific noise estimates, NANO adapts a single model across varying acquisition conditions. The proposed formulation learns the inverse mapping in a lifted latent function space, combining global-local transient representations with stochastic frequency encoding and physics-guided reconstruction. For a fixed discretization and finite latent depth, we establish Lipschitz stability with respect to perturbations in both the measurement and the estimated noise parameter. Under an additional contractivity condition, the associated fixed-point iteration converges to a unique equilibrium. The empirical sensitivity measurements show limited perturbation amplification at the sampled latent states along the tested directions. 
Experiments demonstrate improved albedo reconstruction
and competitive depth accuracy on simulated data,
together with clearer structures in the evaluated
real measurements.
The same trained model also handles reduced exposure
and spatial sampling densities after preprocessing
to the reference grid.
However, under severe spatial downsampling, as depicted in Figure~\ref{downsample-bike}, some fine-scale boundaries remain less completely resolved, suggesting a potential trade-off between noise robustness and local structural fidelity. Future work will explore wave-based neural operators that incorporate virtual-wave propagation to explicitly model diffraction and frequency-dependent propagation effects for superior boundary recovery. 

\funding{The work is supported by the National Natural Science Foundation of China (NSFC 12671658 and T2541053).}

\data{Data underlying the results presented in this paper are available in Ref. \cite{chen2020learned}, Ref. \cite{chang2015shapenet}, Ref. \cite{lindell2019wave} and Ref. \cite{li2023nlost}.}

\appendix

\section{Proof of Theorem \ref{stable-analysis}}
\label{stable}

\begin{proof}
Throughout the proof, the discretization and trained
parameters are fixed.  For two admissible inputs
$(\tau,s_\sigma),(\tau',s'_\sigma)\in\mathcal D$, define
\[
    \delta_\tau
    :=
    \|\tau-\tau'\|_{\mathcal X},
    \qquad
    \delta_s
    :=
    |s_\sigma-s'_\sigma|.
\]
For an affine mapping \(F(x)=Wx+b\) and an \(L_\varphi\)-Lipschitz activation \(\varphi\),
$$
\|F(x)-F(x')\|
\leq
\|W\|\|x-x'\|,
\qquad
\|\varphi(x)-\varphi(x')\|
\leq
L_\varphi\|x-x'\|.
$$
Moreover,
$$
\|F(x)\|
\leq
\|W\|\|x\|+\|b\|,
\qquad
\|\varphi(x)\|
\leq
L_\varphi\|x\|+\|\varphi(0)\|.
$$
The fixed resampling and Fourier mappings, together with
restriction and prolongation, are bounded affine
mappings. Additive skip connections preserve finite Lipschitz
bounds, since
\[
\|v+Tw-(v'+Tw')\|
\leq
\|v-v'\|+\|T\|\|w-w'\|.
\]
With the SFE realization fixed, the SFE module is also
Lipschitz on bounded feature sets.
The transient and noise lifting branches are finite
compositions of affine maps and Lipschitz activations.
They are therefore Lipschitz continuous and map
bounded sets to bounded sets. Hence, there exist finite constants $L_\tau,L_s\geq0$ such that
\begin{equation}
\label{eq:lifting-difference}
    \|\widehat\tau-\widehat\tau'\|
    \leq
    L_\tau\delta_\tau,
    \qquad
    \|\widehat s_\sigma-\widehat s'_\sigma\|
    \leq
    L_s\delta_s.
\end{equation}


Let \( C_{\mathrm{l}}\) denote the Lipschitz constant of the channel-matching mapping used only at initialization. Since the discrete LCT operator \(R_{\mathrm{LCT}}\) is bounded, we have
\begin{equation*}
|\widehat u^0-\widehat u^{\prime0}|
\leq
C_{\mathrm l}\|R_{\mathrm{LCT}}\|L_\tau\delta_\tau.
\end{equation*}
Define $L_{g,0}
:=
C_{\mathrm l}
\|R_{\mathrm{LCT}}\|
L_\tau.$
The mappings used to construct the conditioning
features are Lipschitz continuous and preserve boundedness. Consequently, with the scale index suppressed, the conditioning features associated with each refinement stage satisfy
\begin{equation}
\label{eq:conditioning-difference}
|\widehat u_g^k-\widehat u_g^{\prime k}|
\leq
L_{g,k}\delta_\tau,
\qquad
|\widehat s_\sigma^k-\widehat s_\sigma^{\prime k}|
\leq
L_{s,k}\delta_s
\end{equation}
for finite constants \(L_{g,k}\) and \(L_{s,k}\). When several refinement steps belong to the same scale, the corresponding conditioning features remain unchanged across those steps.

Starting from the bounded initialization, induction through the finite computation graph shows that all attained latent states are uniformly bounded.
For the noise-dependent modulation, write
\[
\mathcal S_{\sigma,k}(w,a;\theta)=q_k(a)\odot w,
\]
with the scale index omitted.
The coefficient mapping \(q_k\) consists of two complex convolutional layers separated by an \(L_\varphi\)-Lipschitz activation. If \(W_{1,k}\) and \(W_{2,k}\) denote the linear parts of these layers, then the biases cancel in differences and
$$
\|q_k(a)-q_k(a')\|_\infty
\leq
L_{q,k}\|a-a'\|,
\qquad
L_{q,k}:=
\|W_{2,k}\|L_\varphi\|W_{1,k}\|.
$$
 On bounded sets of conditioning features and states,
let
\[
\|q_k(a)\|_\infty\leq Q_k,
\qquad
\max\{\|w\|,\|w'\|\}\leq R_k.
\]
Then
\[
\begin{aligned}
\|\mathcal S_{\sigma,k}(w,a;\theta)
-\mathcal S_{\sigma,k}(w',a';\theta)\|
&\leq
\|q_k(a)\odot(w-w')\|
+
\|(q_k(a)-q_k(a'))\odot w'\|\\
&\leq
Q_k\|w-w'\|
+
L_{q,k}R_k\|a-a'\|.
\end{aligned}
\]
Using~\eqref{eq:conditioning-difference}, this gives
\begin{equation*}
\|\mathcal S_{\sigma,k}
       (\widehat u^k,\widehat s_\sigma^k;\theta)
      -
      \mathcal S_{\sigma,k}
       (\widehat u'^k,\widehat s_\sigma^{\prime k};\theta)\|\leq
 Q_k\|\widehat u^k-\widehat u'^k\|
 +
 L_{q,k}R_kL_{s,k}\delta_s.
\end{equation*}
Let $L_{\mathrm{sc},k}$ denote a global Lipschitz
bound for the corresponding spectral block.
Spectral normalization controls the layer norms
entering this bound; non-expansiveness of the
complete block is not assumed.
The refinement update satisfies
\begin{align*}
\|\widehat u^{k+1}-\widehat u^{\prime k+1}\|
&\leq
\|\widehat u^k-\widehat u^{\prime k}\|
+
|\xi_k(\theta)|L_{\mathrm{sc},k}
\Big(
\|\widehat u_g^k-\widehat u_g^{\prime k}\|
+
\|\mathcal S_{\sigma,k}
(\widehat u^k,\widehat s_\sigma^k;\theta)
-
\mathcal S_{\sigma,k}
(\widehat u^{\prime k},\widehat s_\sigma^{\prime k};\theta)\|
\Big)
\nonumber\\
&\leq
A_k\|\widehat u^k-\widehat u^{\prime k}\|
+B_k\delta_\tau+E_k\delta_s,
\label{eq:nonhomogeneous-stage-estimate}
\end{align*}
where
\[
A_k:=1+|\xi_k(\theta)|L_{\mathrm{sc},k}Q_k,~~~
B_k:=|\xi_k(\theta)|L_{\mathrm{sc},k}L_{g,k},~~~
E_k:=|\xi_k(\theta)|L_{\mathrm{sc},k}
L_{q,k}R_kL_{s,k}.
\]
These constants are finite and uniform over
the admissible input set at each refinement stage.

Propagating these estimates through the complete
finite computation graph yields
\[
\|\widehat u^K-\widehat u^{\prime K}\|
\leq
L_{\Phi,\tau}\delta_\tau
+
L_{\Phi,s}\delta_s.
\]
The constants include the initialization bounds,
the refinement coefficients $A_k,B_k,E_k$, and
the bounds for restriction, prolongation,
skip connections, and fixed SFE.
They are finite and uniform over the admissible
input set.

The LCT anchor depends only on the measurement
and satisfies
\[
\|\widetilde u_g-\widetilde u_g'\|
\leq L_g\delta_\tau.
\]
The physics-anchored reconstruction therefore gives
\[
\|\widetilde u^K-\widetilde u^{\prime K}\|
\leq
\mu L_g\delta_\tau
+
C_{\mathrm a}
\|\widehat u^K-\widehat u^{\prime K}\|,
\]
where $C_{\mathrm a}
=
\bigl(1-\mu(\theta)\bigr)
\operatorname{Lip}
\bigl(\operatorname{Conv}_{\mathbb C}
(\,\cdot\,;\theta)\bigr)$. 
Let \(C_{\mathbf p}\) denote a Lipschitz bound for
the subsequent volume projection. It follows that
\[
\|\mathcal G_\sigma(\tau,s_\sigma;\theta)
-\mathcal G_\sigma(\tau',s_\sigma';\theta)\|_{\mathcal U}
\leq
C_{\tau,K}\delta_\tau+C_{s,K}\delta_s,
\]
with $C_{\tau,K}
=
C_{\mathbf p}
\bigl(\mu(\theta)L_g+C_{\mathrm a}L_{\Phi,\tau}\bigr)$ and $C_{s,K}
=
C_{\mathbf p}C_{\mathrm a}L_{\Phi,s}$. 
All constants are uniform over \(\mathcal D\) and
depend only on the admissible set and the fixed network realization.
\end{proof}

\section{Network Architecture Details}\label{detail-net}
Comprehensive detail of the network architecture is summarized below.
\begin{table*}[ht]
\centering
\caption{Network configurations of the proposed NANO framework.}
\label{tab:nano_architecture}
\renewcommand{\arraystretch}{1.}
\setlength{\tabcolsep}{5pt}
\begin{tabular}{
>{\centering\arraybackslash}m{1cm}
>{\centering\arraybackslash}m{2cm}
>{\raggedright\arraybackslash}m{10cm}}
\hline
\textbf{Stage} & \textbf{Module} & \textbf{Configuration} \\
\hline
NLE & Noise estimation &
A 3D U-Net backbone consisting of a five-layer encoder with $3^3$ kernels, strides $(1,2,2,2,2)$, channel sequence $1\rightarrow8\rightarrow16\rightarrow32\rightarrow64\rightarrow128$ and skip connections to a symmetric decoder. The bottleneck features are globally pooled and linearly projected to $\sigma$ and $D_n$, yielding $s_\sigma$ defined as Eq.~\ref{eq:noiselevel};\\ \hline
\multirow{7}{*}{NANO} & \multirow{4}{*}{$\mathcal{B}_{l}$} &\textbf{Noise-level lifting:} The scalar is mapped to a 24-channel tensor via a two-layer MLP\\
\cline{3-3}
&& \textbf{G-TL:} Four-level 3D U-Net with $3^3$ kernels, initial stride $2$, encoder channels $1\rightarrow4\rightarrow8\rightarrow16\rightarrow32$, and symmetric decoder to $4$ channels
\\
\cline{3-3}
& &\textbf{L-TL:} $3^3$ transient downsampling with stride $2$; $3\times3$ spatial windowing; three-level temporal 1D U-Net with $3$-tap kernels and channels $36\rightarrow72\rightarrow144$; MLP $4\rightarrow16\rightarrow4$
\\ \cline{3-3}
& & \textbf{GL fusion:} Two $3^3$ convolutions with channels $8\rightarrow32\rightarrow4$, followed by an FFT;\\
\cline{2-3}
& \multirow{2}{*}{$\Phi$} &
\textbf{Multiscale spectral operator:} Channel matching: Parallel complex 3D convolutions ($5^3$ and $3^3$ kernels), followed by a $1\times1\times1$ complex projection to 24 latent channels;
Spectral network: A five-scale complex spectral network ($24\rightarrow48\rightarrow72\rightarrow96\rightarrow120$ channels). Each scale features two spectral layers ($3^3$ kernels, stride 1 and padding 1), with inter-scale restriction and prolongation using $2^3$ kernels and stride 2.
\\ \cline{3-3} & &\textbf{SFE:} Stochastic frequency encoding applied at the coarsest scale
\\  \cline{3-3} &  
&\textbf{$\textbf{Conv}_{\mathbb C}$:} Complex $1^3$ convolution with channels
$24\rightarrow4$;
learnable weighted fusion with the LCT physics anchor;\\ \hline
VPL & Volume projection &
Depth-wise max/argmax projection;
two parallel 2D rendering heads for intensity and depth, each using $3^2$ convolutions and four convolution blocks;
channels $8\rightarrow8$ and $9\rightarrow16\rightarrow1$; $\times2$ bilinear upsampling \\
\hline
\end{tabular}
\end{table*}
%
%

\bibliographystyle{unsrt}
\bibliography{reference}

\end{document}